%% file: main.tex
\documentclass[screen,acmtog]{acmart}

\usepackage{graphicx}
\usepackage{tikz}
\usepackage{calc}
\usepackage[normalem]{ulem} 
\usepackage{placeins}
\usepackage{array} 
\usepackage{multirow}
\usepackage{hhline} 

\input{macros.tex}
\input{figures.tex}

\AtBeginDocument{%
  \providecommand\BibTeX{{%
    \normalfont B\kern-0.5em{\scshape i\kern-0.25em b}\kern-0.8em\TeX}}}

\setcopyright{acmcopyright}\acmJournal{TOG}
\acmYear{2022}\acmVolume{41}\acmNumber{4}\acmArticle{72}\acmMonth{7} \acmDOI{10.1145/3528223.3530170}

\acmSubmissionID{677}

\begin{document}

\title{Disentangling Random and Cyclic Effects in Time-Lapse Sequences}

\author{Erik H{\"a}rk{\"o}nen}
\authornote{Part of work done during an internship with NVIDIA Research.}
\affiliation{%
  \institution{Aalto University}
  \country{Finland}
}

\author{Miika Aittala}
\affiliation{%
  \institution{NVIDIA}
  \country{Finland}}

\author{Tuomas Kynk{\"a}{\"a}nniemi}
\affiliation{%
  \institution{Aalto University}
  \country{Finland}}

\author{Samuli Laine}
\affiliation{%
  \institution{NVIDIA}
  \country{Finland}}

\author{Timo Aila}
\affiliation{%
  \institution{NVIDIA}
  \country{Finland}}

\author{Jaakko Lehtinen}
\affiliation{%
  \institution{Aalto University \& NVIDIA}
  \country{Finland}}


\renewcommand{\shortauthors}{H{\"a}rk{\"o}nen, et al.}

\begin{abstract}
Time-lapse image sequences offer visually compelling insights into dynamic processes that are too slow to observe in real time. However, playing a long time-lapse sequence back as a video often results in distracting flicker due to random effects, such as weather, as well as cyclic effects, such as the day-night cycle. We introduce the problem of disentangling time-lapse sequences in a way that allows separate, after-the-fact control of overall trends, cyclic effects, and random effects in the images, and describe a technique based on data-driven generative models that achieves this goal. This enables us to ``re-render'' the sequences in ways that would not be possible with the input images alone. For example, we can stabilize a long sequence to focus on plant growth over many months, under selectable, consistent weather.

Our approach is based on Generative Adversarial Networks (GAN) that are conditioned with the time coordinate of the time-lapse sequence. Our architecture and training procedure are designed so that the networks learn to model random variations, such as weather, using the GAN's latent space, and to disentangle overall trends and cyclic variations by feeding the conditioning time label to the model using Fourier features with specific frequencies. 

  
We show that our models are robust to defects in the training data, enabling us to amend some of the practical difficulties in capturing long time-lapse sequences, such as temporary occlusions, uneven frame spacing, and missing frames.
\end{abstract}

\ccsdesc[500]{Computing methodologies~Machine learning}
\ccsdesc[500]{Computing methodologies~Learning latent representations}
\ccsdesc[300]{Computing methodologies~Image representations}
\ccsdesc[300]{Computing methodologies~Computer graphics}

\keywords{Generative Adversarial Networks, controllability, interpretability}

\figTeaser{fig:teaser}

\maketitle

\ifx\confnotice\undefined
\else
    \confnotice
\fi


\section{Introduction}

Time-lapse image sequences depict a fixed scene over a long period of time, enabling compelling visualization of dynamic processes that are too slow to observe in real time. Such processes include both natural phenomena like plant growth, weather formation, seasonal changes, and melting glaciers, as well as human efforts like construction and deforestation. Unless shot in highly controlled settings, time-lapse sequences mix together deterministic, often cyclic effects over different time scales, as well as more random effects, such as weather, traffic, and so on. When playing the sequence back as video, these effects typically result in distracting flicker.

Given an input time-lapse sequence, our goal is to disentangle the appearance changes due to random variations, cyclic fluctuations, and overall trends, and to enable separate control over them. Concretely this means, for example, that a raw sequence that shows plant growth over many months, with the images featuring different times of day as well as different weather conditions, can be ``re-rendered'' so that the weather and time of day remain approximately fixed --- at whatever we choose from among those featured in the input sequence --- and the plant growth remains the only visible factor. This enables not only pleasing visualizations, but also helps gain better insight into the depicted phenomena. See Figure~\ref{fig:teaser}.

We believe we are the first to study this problem. Prior work aims to either hallucinate a time-lapse from a single input photo, to generate novel plausible time-lapses that feature scenes not seen during training \cite{Endo2019TimelapseWarp,Logacheva2020DeepLandscape}, or to stabilize a given sequence into a temporally smooth version without control over the different \mbox{factors of variation \cite{MartinBrualla2015TimelapseMining}.}

We build on modern data-driven generative models' ability to generate realistic, high-resolution images. Specifically, we train, for each individual input sequence, a separate StyleGAN2 \cite{Karras2020stylegan2} generator that takes the time coordinate along the sequence as a conditioning variable. Our networks learn to model random variations, such as weather, using the GAN's latent space, and to model deterministic variations by the conditioning time variable. To further enable disentanglement and separate control over both overall trends and cyclic effects such as day-night cycles and seasonal changes, we use Fourier features with carefully chosen frequencies computed from the input timestamp.

We demonstrate our method on several time-lapse datasets from the AMOS collection \cite{Jacobs07AMOS}, with sequence lengths typically between 2 and 6 years, as well as a proprietary plant growth dataset collected over a summer.


In the figures, we visualize time-lapse sequences as \emph{time-lapse images} where time passes from left to right (Figure~\ref{fig:timelapse_image}). We construct these images as follows. For each $x$-coordinate of the time-lapse image, we first assign an interpolated time value $t_x$ within the time-lapse sequence. We then populate the image by copying vertical columns of pixels, indexing the source sequence using $t_x$ and $x$. For example, if the time period is one year, this will visualize all seasons in the same image.

Our contributions are summarized as follows:
\begin{itemize}
    \item We present the first technique for disentangling appearance changes due to trends, cylic variations, and random factors in time-lapse sequences.
    \item We propose a new conditioning mechanism for GANs that is suitable for learning repeating, cyclic changes.
\end{itemize}
\radd{Our implementation, pre-trained models, and dataset preprocessing pipeline are available at \url{https://github.com/harskish/tlgan}}.

\figTimelapseImage{fig:timelapse_image}


\section{Related Work}

Martin-Brualla et al.~\shortcite{MartinBrualla2015TimelapseMining} describe an optimization-based method for removing flickering in time-lapse sequences by temporally regularizing the $\text{L}_1$ or $\text{L}_2$ difference between two consecutive time-lapse frames, on a per pixel basis. While their method is effective at removing flickering, it leads to undesired blurring in the resulting time-lapse sequence if the geometry of the images changes significantly during the sequence, e.g., if trees are growing or new buildings are built in the scene. Moreover, their method is “playback” only, meaning it can be used to compile a single time-lapse video, whereas our method offers the user a control over cyclic and random effects within the sequence.

Deep generative models, including generative adversarial networks (GAN) \cite{Goodfellow2014GAN}, variational autoencoders (VAE) \cite{Kingma2014VAE}, autoregressive models \cite{VanDenOord2016APixelCNN,VanDenOord2016BPixelRNN}, flow-based models \cite{Dinh2016RealNVP,Kingma2018Glow}, and diffusion models \cite{SohlDickstein2015Diffusion,Yang2019DataGradients,Ho2020DDPM}, have been used to reach impressive results in wide variety of tasks, including image synthesis \cite{Karras2019StyleGAN,Brock2019BigGAN,Karras2020stylegan2,Karras2020ADA,Karras2021AliasFree}, image-to-image translation \cite{Zhu2017CycleGAN,Choi2020StarGAN2,Kim2020UGATIT}, and controllable editing of still images \cite{Park2019SPADE,Karacan2019GauGANish,Park2020SwappingAE,Huang2021PoEGAN}.
They have also been used in video synthesis \cite{Clark2019DVDGAN,Wang2018Vid2vid,Tulyakov2018MoCoGAN,Wang2019FewshotVid2vid}.
%

Recently, many generative models that \emph{hallucinate} a short time-lapse sequence from a single starting image have been proposed \cite{Xiong2018MSDGAN,Daichi2020SSAGAN,Colton2021GANlapse,Endo2019TimelapseWarp,Logacheva2020DeepLandscape}. These models are trained with a large database of time-lapse videos. 
Among the proposed methods, Logacheva et al.~\shortcite{Logacheva2020DeepLandscape} is the closest to our work. They modify the original StyleGAN network \cite{Karras2019StyleGAN} such that it includes latent variables that can be used to separately control static and dynamic aspects in the images. Time-lapse from a single input image is generated by first embedding it to the latent space of the generator network, and then varying the latent variables that control the dynamic aspects.


Another perspective to hallucinating time-lapse videos is to apply image-to-image translation \cite{Nam2019VideoSynthesis,Anokhin2020HiDT}.
Nam et al.~\shortcite{Nam2019VideoSynthesis} use conditional GANs, conditioned on timestamps from one day, to synthesize a time-lapse from a single image. Their method can change the illumination of the input, based on the conditioning time-of-day signal, but it cannot synthesize motion, and thus moving objects, such as clouds, appear fixed.
Anokhin et al.~\shortcite{Anokhin2020HiDT} modify the lighting conditions of a target image sequence by translating the style from a separate source sequence. Their model can produce plausible lighting changes but as it works on a frame-by-frame basis it leads to flickering of \mbox{moving objects in the scene.}

Our goal is not to invent time-lapse videos based on an image, but rather to \emph{process actual time-lapse sequences} so that different effects can be disentangled, and the output can be stabilized and controlled in a principled way.
To the best of our knowledge, generative models have not been used for
this purpose before.


\section{Probabilistic Generative Modeling of Time-lapse Sequences}
\label{sec:timelapses}

The appearance of a natural scene over long periods of time typically features random components intermixed with deterministic effects. For instance, the day-night cycle is entirely deterministic, whereas the weather (rainy or clear) may vary randomly; the time of year may alter the probability of rain, cloudiness, or snow cover at a given time of day; and finally, long-time trends may show growth of trees, construction of buildings, etc. 

We seek a generative model that produces random images that \emph{could} have been taken at a specified time of day \radd{$c_d$}\rdel{$t_d$}, day of year \radd{$c_y$}\rdel{$t_y$}, and global trend \radd{$c_g$}\rdel{$t_g$}. To this end, we interpret the frames in a time-lapse sequence as samples from the conditional distributions
\begin{equation}
p(\text{image}\; |\; c_d, c_y, c_g) \label{eq:conditional_distribution}
\end{equation}

and use it to train a generative model $G(\bz;\; c_d, c_y, c_g)$, a function that turns Gaussian random latent variables $\bz \in \mathcal{N}(\boldsymbol{0}, \boldsymbol{I})$ into images. While images in the training set come with fixed combinations of $(c_d, c_y, c_g)$, a successful model will learn a disentangled representation that allows them to be controlled independently, and pushes the inherent random variation into the latent space. This allows previously unseen applications, such as exploring random variations at a particular fixed time, or stabilizing the appearance by fixing the latent code and only showing variations over different timescales.\footnote{The reader may notice that the inclusion of the global trend $t_g$ collapses the conditional distribution \eqref{eq:conditional_distribution} to a Dirac impulse, i.e., there is only a single training image consistent with each combination of $(c_d, c_y, c_g)$ in the training data. This means our generative modeling problem is, in this basic form, ill-posed and, in principle, solvable by memorization. In the following section, we describe label jitterings techniques that remove this shortcoming.} 

\figDefects{fig:defects}

The conditional distribution view to time-lapse sequences reveals why a regression model that would deterministically learn to output a given image at a given $(c_d, c_y, c_g)$ is not able to learn the disentangled representation we seek: even if equipped with cyclic time inputs, the model is required to reproduce the \emph{particular} images at \emph{particular} time instants, i.e., to bake random effects together in with the time inputs. This would lead to a lack of disentanglement, and in practice, blurry results due to finite model capacity.

Our formulation encourages the model to attribute repeating patterns in the appearance distributions to the cyclic input variables $c_d$ and $c_y$. This has the great benefit that training is robust to the significant gaps often found in long time-lapse datasets (Figure~\ref{fig:defects}a): as long as there are other days or years where images from the missing combination can be found, our models are able to hallucinate plausible content to the missing pieces.

In addition to random changes in the scene, real time-lapse sequences feature several other common types of defects that further increase apparent randomness and underline the need of a probabilistic model. These include changes in alignment and camera parameters during servicing (Figure~\ref{fig:defects}b), camera hardware updates, and due to thermal expansion and contraction; 
adaptive ISO settings based on scene brightness; and
temporary occlusions, e.g., spider webs, condensation, and ice.


\section{Architecture and training}

\figArchitecture{fig:architecture}

We build on the StyleGAN2 \cite{Karras2020stylegan2} model that has achieved remarkable results in synthesizing realistic high-resolution images. In this section, we describe how the cyclic conditioning signals are provided for the generator and discriminator. 
We also describe how the model is trained.

\subsection{Architecture}

\paragraph{StyleGAN2 Architecture.}
The distinguishing feature of StyleGAN2 is its unconventional generator architecture, Figure~\ref{fig:architecture}. Instead of feeding the input latent code $\zz \in \ZZ$ only to the beginning of the network, the \emph{mapping network} first transforms it to an intermediate latent code $\ww \in \WW$.
Affine transforms then produce \emph{styles} that control the layers of the \emph{synthesis network} by modulating convolution weights. 
Additionally, stochastic variation is facilitated by providing additional random noise maps to the synthesis network.


\paragraph{Our Architecture.}

We encode the conditioning signals as follows:
\begin{equation}
    \label{eq:conditioning}
    \mathbf{c}(t_d, t_y, t_g) = \begin{bmatrix} 
        \mathbf{c}_d(t_d) \\
        \mathbf{c}_y(t_y) \\
        \mathbf{c}_g(t_g)
    \end{bmatrix}
    = \begin{bmatrix} \sin(2 \pi f_0 t_d) \\ \cos(2 \pi f_0 t_d) \\ \sin(2 \pi f_1 t_y) \\ \cos(2 \pi f_1 t_y) \\ t_g \radd{\cdot k} \\ 1 \end{bmatrix},
\end{equation}
%
where $f_0$ matches the day cycle, $f_1 = f_0 / 365.25$ matches the year cycle, \radd{and the scaling constant $k=1 \times 10^{-2}$ makes the relative learning rate of the trend component smaller}. \radd{During training, the input linear timestamps $t_d=t_y=t_g \in [0,1]$ are identical and normalized across the whole time-lapse sequence, with 0 corresponding to the time and date of the first image and 1 to those of the last image. After training, they can be modified independently to change only certain aspects of the output image.}
This mechanism has similarities with Fourier features \cite{Tancik2020FourierFeatures} and positional encoding of transformers \cite{Vaswani2017Attention}, as they both use stacks of sinusoids to map from lower to higher dimensional space.

As illustrated in Figure \ref{fig:architecture}, we feed the conditioning signals directly to each layer of the synthesis network and provide a simple mechanism for them to control the styles. 
Similar direct manipulation of styles has previously been shown to yield excellent disentanglement in the editing of GAN-generated images \cite{Collins2020,Kafri2021,Chong2021}.
For each layer $i$, the conditioning signals are transformed by a learned linear transformation $\mathbf{L}_i$ into a scale vector $\mathbf{k}_i = \mathbf{L}_i\mathbf{c}$ with the same dimensionality as the corresponding style vector $\mathbf{s}_i$. This scale vector is then used to modulate the style vector by element-wise (Hadamard) multiplication, i.e., $\mathbf{s}'_i = \mathbf{k}_i \odot \mathbf{s}_i$, and the resulting scaled style vector $\mathbf{s}'_i$ replaces the original style vector $\mathbf{s}_i$ when modulating the convolution kernels. 
We also tried introducing another set of linear transforms to produce $\mathbf{c}$-dependent biases for each layer. This allowed the conditioning to manipulate styles also in an additive way, but it did not lead to further improvement.

Intuitively, the linear transformations $\mathbf{L}_i$ are able use the sinusoidal inputs to build detailed, time-varying scaling factors for each feature map in the layer. Since $\mathbf{c}$ contains both sine and cosine parts of the cyclic signals as well as constant and linear terms, the output linear combinations can contain cyclic signals with arbitrary phases, offsets, and linear trends, including purely linear or constant scale factors.
As an example, Figure~\ref{fig:day_length} showcases more complex behavior where the timing of sunrise and sunset depends on both time-of-day ($c_d$) and time-of-year ($c_y$) simultaneously.

For the discriminator, we adopt the approach of Miyato and Koyama \shortcite{Miyato2018ProjD} by evaluating the final discriminator output as $D(x) = \mathrm{normalize}(M(c)) \cdot D'(x)$, where $D'(x)$ corresponds to the feature vector produced by the last layer of the discriminator. $M(c)$ represents a learned embedding of the conditioning vector that we compute using using a dedicated 8-layer MLP.\footnote{Note that this is the same mechanism employed by the official implementation of StyleGAN2-ADA \cite{Karras2020ADA}.}

\figDayLength{fig:day_length}

\subsection{Training}

We follow the general procedure for training a conditional GAN model: the generator and the discriminator are trained simultaneously, and the conditioning labels of the generated and real images are also passed to the discriminator. The conditioning labels are sampled from the training set, and augmented with noise as described below. We use the StyleGAN2-ADA training setup \cite{Karras2020ADA} in the 'auto' configuration, keeping most of the hyperparameters at their default values. In practice, we have found it beneficial to increase the R1 gamma slightly: we use values 4.0 and 16.0 for $512 \times 512$ and $1024 \times 1024$ models, respectively. We  use batch size 32 for all datasets.


\tabSceneDetails{tab:scene_details}

The goal is to train the generator so that any input timestamp generates a reasonable distribution of output images, even if it is missing from the training data. The timestamps in the data suffer from several subtle issues related to regular discrete sampling, missing intervals, and the theoretical possibility of simply memorizing the timestamp-to-image mapping. In the following, we introduce two timestamp jittering mechanisms to eliminate these problems.

\subsubsection{Timestamp dequantization}
\label{sec:dequantization}
Whereas our training sequences contain photographs taken at specific times --- often at regular intervals, such as at every 30 minutes --- our goal is to create models that can be evaluated at arbitrary points. To this end, we add noise to the input labels with the goal of mapping every continuous time value to a valid training image. We call this \emph{dequantization}. It is applied to the inputs of both the generator and discriminator during training.

The $j$'th image in the temporally ordered dataset is associated with a raw linear timestamp $T_j \in [0,1]$\radd{, which is used to compute the corresponding conditioning triplet $(\mathbf{c}_d, \mathbf{c}_y, \mathbf{c}_g)$}. 
Whenever a timestamp $T_j$ is used in training (either to condition the generator, or when sampling real images for the discriminator), we first jitter it by a random offset $\epsilon \sim \mathcal{N}(0, \max(T_{j+1}-T_j, T_j-T_{j-1})/2)$ that is proportional to the temporal distance to its neighboring frames. This spreads the timestamps of existing images to fill any gaps in the dataset.

When a dataset systematically lacks, e.g., night-time images, our label noise basically redirects the missing timestamps to the nearest morning and evening images. Obviously, night-time images cannot be synthesized if they were never seen in training.
With random gaps the situation is different, and we can indeed learn to fill the gaps using plausible variation. Consider a multi-year dataset where all data from July 2015 is missing. The label noise again fills this gap with nearby neighbors. Now, when training the model, we sample all training images with equal probability. If we have plenty of samples from July 2014 and 2016, those will be used frequently in training, while the 2 neighbors of July 2015 will be sampled only rarely. The time-of-year signal thus learns to essentially ignore the gap and fill it using the other years. The same is true for all time scales.



\subsubsection{Discriminator timestamp augmentation}
\label{sec:timestampnoise}

%
As detailed in Section~\ref{sec:timelapses}, the conditional distributions of Equation~\eqref{eq:conditional_distribution} are almost degenerate, i.e., each input tuple $(c_d, c_y, c_g)$ is only associated with a small number of training images even under the label dequantization scheme. To combat this and encourage the model to share information between similar conditions, we build on the intuition that the distribution of images taken at, say, 12:00 noon on March 15 should not look too different from images taken around the same time on March 13 or March 20, and that the global trend should only pay attention to effects clearly longer than a year.

We implement this by adding, in addition to the dequantization noise described above, independent noise to the raw linear timestamps $t_y$ and $t_g$ for both real and generated images upon passing them to the discriminator. Specifically, we add Gaussian noise of $\sigma_y = \text{1 week}$ to the time of year input $t_y$, and noise of $\sigma_g \in \{1.5\text{ years}, 2\text{ years}\}$ to the global trend input $t_g$ depending on the dataset, see Table~\ref{tab:scene_details}. The former makes it impossible for the model to discern the precise day within the year, and the latter makes the global component $t_g$ essentially uncorrelated with the other inputs, making it impossible for the model to capture anything but effects of the longest time scales using $t_g$.
This process can also be seen to ``inflate'' or augment the training data so that a single moment in time corresponds to a much larger set of possible images, each of which is still (roughly) consistent with the given time-of-year and time-of-day. In our tests, this makes convergence more reliable also in smaller datasets.
The noise never seems to hurt larger datasets, although in such cases we observe that the inductive biases of our architecture guide the learning to a similar disentangled representation even without the addition of conditioning noise -- section~\ref{sec:training_behavior} discusses this further.


\section{Results and Comparisons}
\label{sec:results}
\figLatents{fig:latents}
\figTimeOfDay{fig:time_of_day}
\figTide{fig:tide}

We will now present example results from our model, and study how the different control mechanisms (latent code, time-conditioning signals) affect the synthesized time-lapses. 
Most of the results are best appreciated from the accompanying videos.

We use 8 datasets from the Archive Of Many Outdoor Scenes (AMOS) \cite{Jacobs07AMOS, jacobs09webcamgis}: \Frankfurt{}, \Kuessnacht{}, \Normandy{}, 
\Barn{}, \Muotathal{}, 
\Teton{}, \Twomedicine{}, and \Valley{}. Of these, the first 3 are urban environments, while the rest are landscapes. All datasets are multiple years long (48k -- 82k frames). In addition, we use a proprietary shorter dataset, \MPP{} (``\textsc{Letters to the Editor}''), that depicts an art installation made of growing plants over a single summer (6k frames over 5 months). Due to its shorter length, we only condition \MPP{} using the time-of-day and global trend signals.
The spatial resolution of the image content ranges from $512\times358$ to $1024\times960$, and is slightly different for almost all datasets. The frames are further zero-padded to the closest power of two to produce a square-shaped image for training.
Typical temporal resolution is one frame per half an hour, with some variation within and between the datasets.  
We provide detailed statistics of all datasets in Table~\ref{tab:scene_details}. 

We selected the datasets from the AMOS collection based on image resolution and quality, sequence length, camera alignment stability, and subjective appeal of scene content. These criteria led to the elimination of all data recorded before 2010. 
The selected datasets were aligned during preprocessing (Appendix~\ref{sec:alignment}).

As our goal is to disentangle the effects of a particular dataset,
we train a separate generator for each time-lapse sequence. 
We train each model until \rdel{kernel}\radd{Fréchet} inception distance (\rdel{KID}\radd{FID}) \cite{Heusel2017FID} stops improving and the roles of conditioning inputs have stabilized, which happens around 6M real images. This takes $\sim$60 hours on 4 NVIDIA V100 GPUs at $1024\times1024$ resolution (30 hours at $512\times512$).
Once the model is trained, images can be generated at interactive rates.


\subsection{Latent space}
As hypothesized in Section~\ref{sec:timelapses}, our model should learn to describe random variations (such as weather) of a time-lapse sequence using the latent space.
In Figure~\ref{fig:latents} we verify that this actually happens.
We chose four time stamps from the \Barn{} dataset to cover different seasons (summer, autumn, winter) and times of day (dawn, noon, evening). We also chose four latent codes that appear to match a foggy day, clear day, partly cloudy, and a high-contrast rainy day.

Clearly, the chosen latent code has a very large effect on the output images. We can also see that the same weather type gets plausibly expressed across the time stamps, indicating that we can now stabilize the weather in synthesized time-lapse sequences of arbitrary length by simply selecting one latent code for the entire sequence. This eliminates the significant flickering exhibited in the input sequence.
By selecting a different latent code, we can switch the entire output sequence from, e.g., clear weather to cloudy weather, as demonstrated in the accompanying video.


\subsection{Time conditioning}
We will now inspect the effect of our three time-conditioning signals, by varying each of them in isolation and observing the resulting changes in the output images. 
Figure~\ref{fig:time_of_day}, top half, shows three datasets where we adjust only the time-of-day signal to observe its effect. We see that this signal has learned to represent the day cycle, as we had hoped. Again, we can synthesize the results using different latent codes to specify the weather and other random aspects. 
Figure~\ref{fig:tide} shows an additional test, where the time-of-day signal controls the sunrise and sunset, and also the tides, closely matching the training data.\footnote{In reality, tidal schedules change between days in a complicated way, which our longer-term signals also try to emulate.}
%
As shown in the bottom half of Figure~\ref{fig:time_of_day}, the time-of-year signal learns to similarly control seasons. 

%
Figure~\ref{fig:trend} shows the effect of varying only the trend signal, indicating that non-cyclical effects like plant growth and the construction or renovation of buildings are controlled by it. 
The variance images visualize which pixels are most strongly affected by the trend component. The exact computation of the variances is explained in Appendix~\ref{sec:variance}.

The accompanying video demonstrates that we can synthesize naturally evolving time-lapse sequences at a much higher sampling rate than the training data, indicating that our model has learned a continuous representation of time.

Figure~\ref{fig:robustness} shows an example where significant chunks of input data are missing and some other time periods have been corrupted by overexposure. As this is a multi-year dataset, our model ends up extrapolating the missing data based on the other years, leading to a plausible result.

\figRobustness{fig:robustness}
\figTrend{fig:trend}


\subsection{Training behavior}
\label{sec:training_behavior}

\begin{figure}
    \centering
    \def\svgwidth{\columnwidth}
    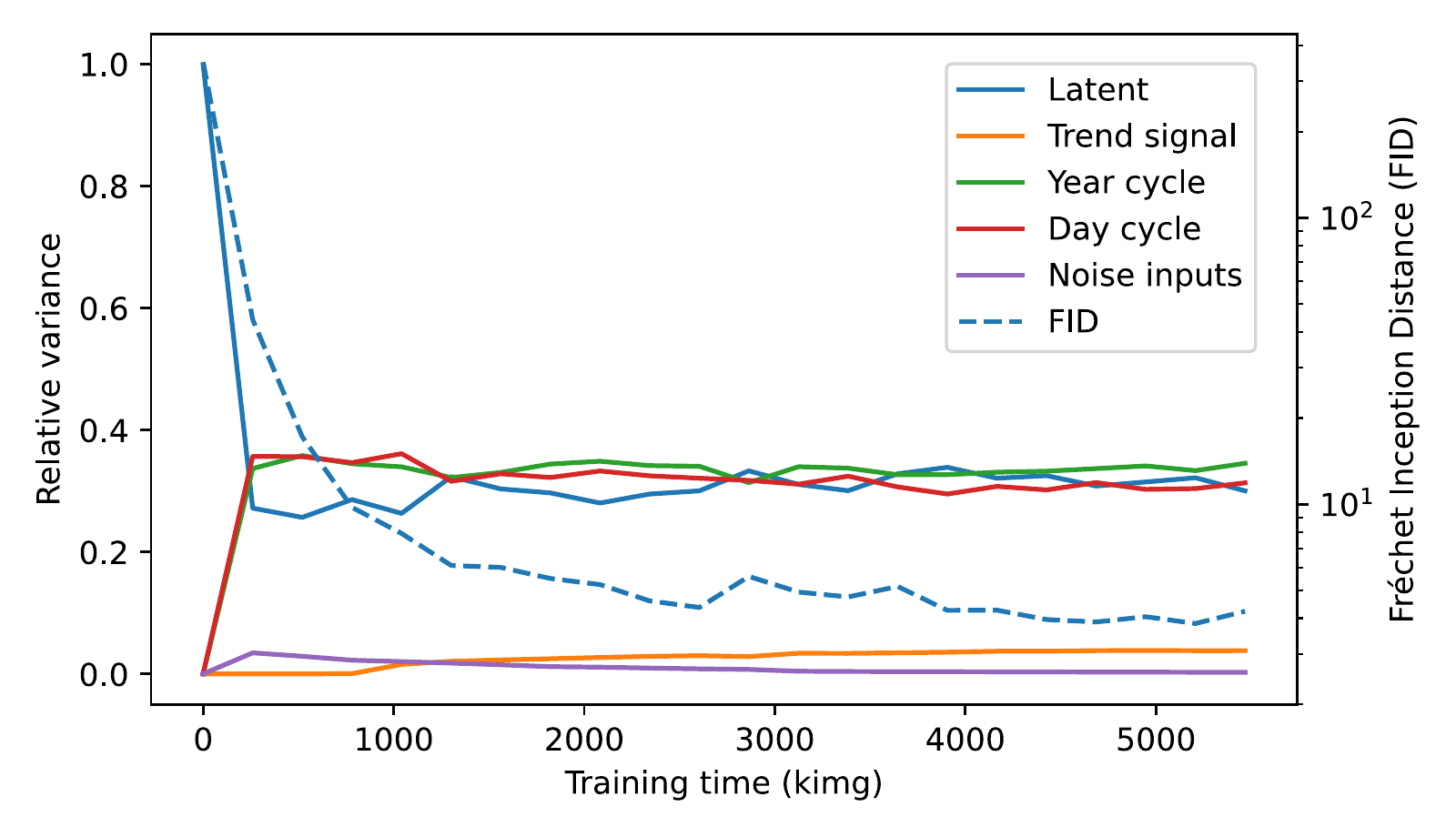
    \caption{Typical convergence behavior for our models (\Valley{} shown): the latent (z) dominates initially, but the day and year cycles quickly start affecting the output. The noise inputs tend to decrease in importance over time, while the linear component typically increases in importance after an initial delay.}
    \label{fig:convergence}
\end{figure}

We use variance (computed as detailed in Appendix~\ref{sec:variance}) to estimate how strongly the synthesized images are affected by different network inputs/control mechanisms: latent code, time-of-day, time-of-year, trend, and StyleGAN2's noise inputs.

Figure~\ref{fig:convergence} plots how the relative variance of components evolves during training in the \Valley{} sequence. The roles of latent code, time-of-day, and time-of-year are learned quickly, and each of them ends up corresponding to $\sim$30\% of the variance in output images. The role of the trend signal is much slower to learn and is ultimately responsible for less than 10\% of the variance, further indicating that the trend is not being abused for memorization. 
The noise inputs of StyleGAN2 have a very minor effect on the output, which implies that the generator has learned to model almost all the variation using the latent code and time conditioning, and basically does not need the additional degrees of freedom offered by the noise inputs.

\begin{figure}
    \centering
    \def\svgwidth{\columnwidth}
    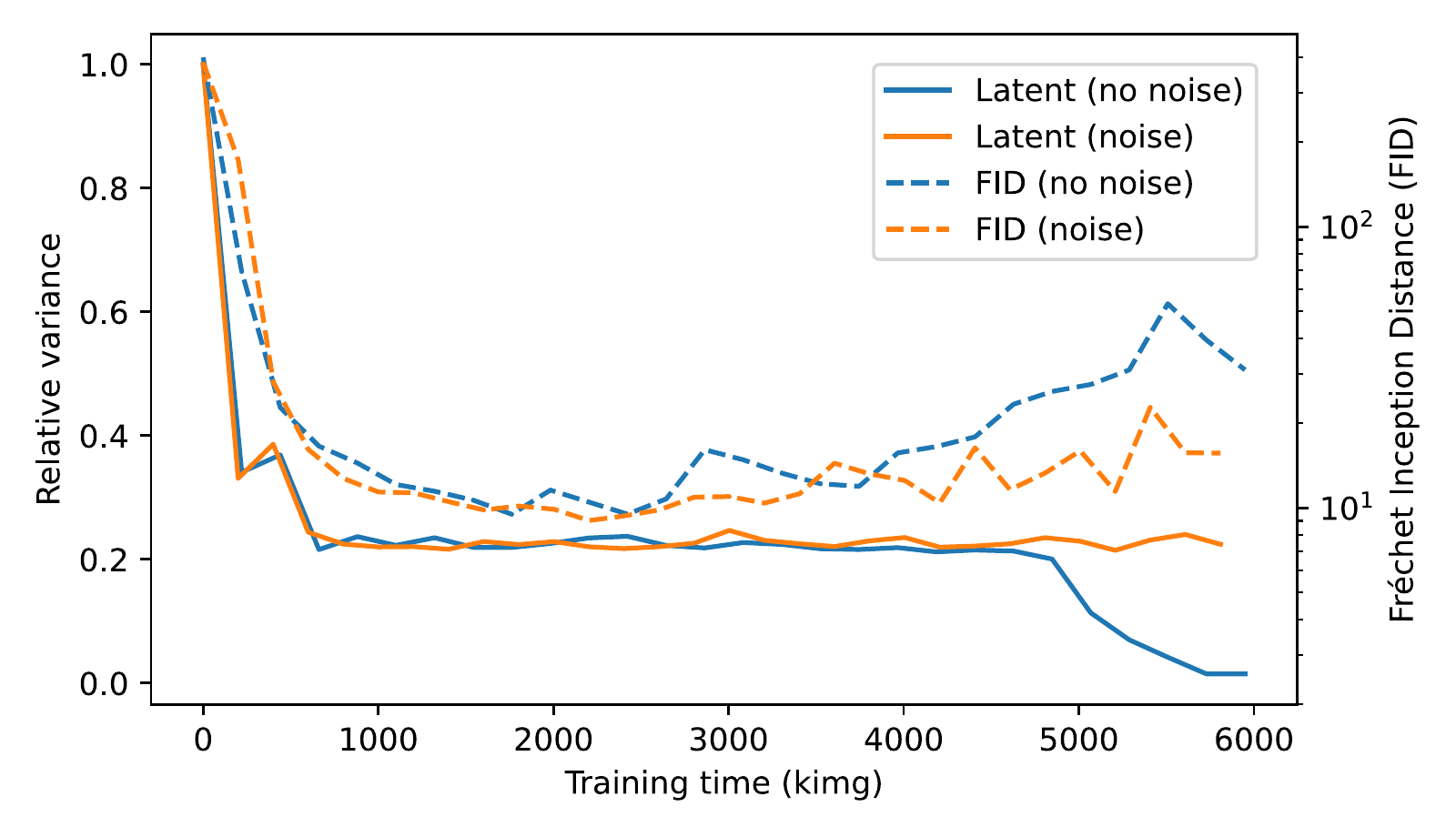
    \caption{\radd{On rare occasions our models can start memorizing the input data, leading to a collapsed state where the time axis is reduced to a few template frames, and all random variation disappears, seen as the variance of the latent approaching zero. We have found that including timestamp noise (Section~\ref{sec:timestampnoise}) and reducing the linear component scale (Equation~\ref{eq:conditioning}) fixes the issue.}
    }
    \label{fig:collapse}
\end{figure}

Figure~\ref{fig:collapse} showcases a rare situation where where the lack of timestamp augmentation noise (Section~\ref{sec:timestampnoise}) causes models trained on smaller datasets to collapse and generate only a few template frames, seen as the variance of the latent approaching zero. Including timestamp noise prevents such failures. Additionally, we find that timestamp augmentation and the scale of the linear component ($k$ in Equation~\ref{eq:conditioning}) interact during training: with a faster adapting, larger-scaled linear component, more noise is needed to prevent memorization.


\figComparison{fig:comparison}
\subsection{Comparisons}
Figure~\ref{fig:comparison} presents a comparison with \cite{MartinBrualla2015TimelapseMining} in \MPP{}. The method produces a single stabilized version of the input sequence by temporal smoothing. As the technique optimizes pixels in isolation --- without regard to how neighboring pixels change together --- it blurs moving content such as growing plants or moving chairs. In contrast, our results are free from such artifacts.
The accompanying video includes the full time-lapse sequences.

\radd{We provide comparisons to \cite{Anokhin2020HiDT} and \cite{Logacheva2020DeepLandscape} in the accompanying video. While both are able to generate subtle changes in lighting, the results clearly show how our models trained on one specific scene are able to generate much more specialized changes, such as shadows being realistically cast by the scene content as the lighting conditions change.}

\figAblationSingleCond{fig:ablation_single_cond}
As an ablation, we trained a variant of our architecture where the cyclical signals ($c_d, c_y$) are left out, and only the global trend ($c_g$) input is included (Ablation A). As expected, this model disentangles the global trend correctly, but all other changes, such as time-of-day and time-of-year, end up in the latent space, giving the user only indirect and imprecise control over the output \radd{-- see Figure~\ref{fig:ablation_single_cond}, top.}

\figAblationFcat{fig:ablation_concat_single}
As a baseline, we trained a simple conditional StyleGAN2 generator, where the (embedded) conditioning signals are concatenated to the latent code $\zz$ \cite{Mirza2014}.
This approach leads to more entangled results than our conditioning architecture.
When only $c_g$ is used (Ablation B), the baseline strongly entangles time-of-year to the global trend \radd{(Figure~\ref{fig:ablation_single_cond}, bottom).}
When also the cyclical signals ($c_d, c_y$) are used (Ablation C), we observe partial entanglement such as the day cycle affecting seasons and trend, and latent $\zz$ affecting the apparent timestamp. \radd{Figure~\ref{fig:ablation_concat_single} showcases one such example, where the day cycle is entangled with the trend, causing a tree to disappear.}


\subsection{Latent exploration}
Several techniques enable post hoc analysis of the features learned by generative models. We use the GANSpace approach \cite{Harkonen2019GANSpace} to discover interpretable directions in the intermediate latent space $\bw$. The first principal component controls overall sunniness in all models, with clear skies, partly cloudy, overcast, and foggy weather being generated in order along the same linear subspace. During wintertime, the amount of snow also changes with the conditions, with cloudy/foggy weather being associated with more snow, and clear sunny weather resulting in more melting. Other typical effects found are controls for the smoothness of bodies of water like rivers and lakes, changes in brightness and saturation, changes in cloud appearance and altitude, and so on. \radd{Interactive exploration of the latent space is showcased in the \mbox{accompanying video}.}



\section{Limitations and future work}
\figIssues{fig:issues}

Clearly, it would be desirable to learn the alignment as a part of the training. One possibility could be to switch to Alias-Free GAN \cite{Karras2021AliasFree}, which includes the concept of input transformations through spatial Fourier Features. 
Currently the alignment issues that remain in the data are learned by the global trend component, as are changes in color temperature or exposure (Figure~\ref{fig:issues}).

As night and day look almost completely different in urban environments (e.g., Figure~\ref{fig:tide}), a generator may handle these separately, in which case there is no guarantee that a latent would yield the same weather for both day and night. 
Also, we find that night-time often has very little variation in the input data, causing the GAN to learn only a few ``templates'' for night-time images.

As our output images are synthesized from scratch using a generator network, some GAN-related artifacts may appear. 
Each frame is generated independently and nothing forces, e.g., the clouds to move plausibly in animation. Explicitly encouraging time-continuity in GAN training is a possible future improvement.
The training of GANs is memory-intensive, and this currently limits the maximum resolution of images to approximately $2048\times2048$. 

We find the quality of the resulting images to be generally good. Close inspection can reveal subtle artifacts, such as grid-like patterns in foliage areas, clouds sometimes being rendered unrealistically, or ringing around strong edges. We interpret these effects as slight signs of collapse, probably caused by the highly correlated nature of the training images. 

\tabFIDcomp{tab:fid_comp_sg2}
\radd{Perhaps somewhat surprisingly, we find that the increased control provided by our method does not come at the cost of image quality -- Table~\ref{tab:fid_comp_sg2}, which contains FIDs of our method and StyleGAN2 \cite{Karras2020stylegan2}, shows no systematic degradation in quality.}

An objective evaluation of disentanglement is difficult due to the lack of generally applicable quantitative metrics. This is an important avenue of future work.

We currently train a separate generator for each dataset. Training on multiple datasets simultaneously and conditioning the model also using the scene ID might offer possibilities for useful transfer between similar-looking datasets. The resolution differences between datasets are a practical hindrance, however.
In targeted tests, we have observed quite reliable disentanglement with as few as 1000 training images. Concurrent training with multiple datasets would likely improve the behavior in the limited-data regime.

Overall, we believe conditional generative models may have more applications in disentangling complex effects in individual datasets.

\begin{acks}
We thank Tero Karras for insightful comments and feedback. \href{http://www.mielipidepalsta.fi/}{\MPP{}} is the work of visual artists \href{https://www.emmaronnholm.com}{Emma Rönnholm} and \href{https://www.sallavapaavuori.net}{Salla Vapaavuori}. We are grateful to them, as well as curators Eerika Malkki and Jari Granholm of the \href{https://purnu.fi/english/}{Purnu Art Center} 2019 summer exhibition Lumous for working with us on capturing the dataset.

This work was partially supported by the European Research Council (ERC Consolidator Grant 866435), and made use of computational resources provided by the Aalto Science-IT project and the Finnish IT Center for Science (CSC).
\end{acks}

\bibliographystyle{apalike}
\bibliography{tlgan}


\appendix
\section{Computation of variance images}
\label{sec:variance}
In order to monitor the training progress and analyze our results, we measure the relative importance of our model inputs, by computing the per-pixel variance of the output with respect to each input.

Given a model $\mathrm{G}$, a random output image $y$ is produced as a function of five input variables:
\begin{equation}
y = \mathrm{G}(\bz, \bn, t_g, t_y, t_d),
\end{equation}
where $\bz \sim \mathcal{N}^{512}(0,1)$ denotes the latent vector, $\bn \sim \mathcal{N}^K(0,1)$ is the tensor of all per-layer noise inputs, $t_g \sim \mathcal{U}(0,1)$ is the global trend timestamp,  $t_y \sim \mathcal{U}(0,1)$ is year cycle timestamp, and $t_d \sim \mathcal{U}(0,1)$ is the day cycle timestamp.

For each of these five parameters, we compute the variance of each output pixel (and color channel) with respect to that parameter, and average this variance over all choices of the remaining parameters. For example, for parameter $z$:
\begin{equation}
V_{\bz} = \mathbb{E}_{\bn, t_g, t_y, t_d} \mathrm{Var}_{\bz} [\mathrm{G}(\bz, \bn, t_g, t_y, t_d)],
\end{equation}
and analogously for the other choices of parameter. An equivalent formulation is amenable for convenient Monte Carlo estimation: 
\begin{equation}
V_{\bz} = \frac{1}{2} \mathbb{E}_{\bz, \bz', \bn, t_g, t_y, t_d} [\mathrm{G}(\bz, \bn, t_g, t_y, t_d) - \mathrm{G}(\bz', \bn, t_g, t_y, t_d)]^2
\end{equation}
In practice, we generate $N=5000$ pairs of images with random parameters, such that within each pair, the relevant parameter (e.g., $z$) is further randomized while the others are held constant. A squared difference is computed for each pair, and the $N$ difference images are averaged.

The five estimated variance images $\tilde V_{\bz}$, $\tilde V_{\bn}$, $\tilde V_{t_g}$, $\tilde V_{t_y}$, and $\tilde V_{t_d}$, are further normalized so as to sum to one at each pixel and color channel. For example, for $\bz$:
\begin{equation}
\tilde V_{\bz}^{\text{norm}} = \frac{\tilde V_{\bz}}{\tilde V_{\bz} + \tilde V_{\bn} + \tilde V_{t_g} + \tilde V_{t_y} + \tilde V_{t_d}}
\end{equation}

Figure~\ref{fig:trend} shows $\tilde V_{t_g}^{\text{norm}}$, averaged over the channel dimension. Figure~\ref{fig:convergence} shows the evolution these normalized variance images, averaged over the channel and the pixel dimensions.


\section{Input Sequence Alignment}
\label{sec:alignment}
Our models are keen to pick up on small changes in image alignment, and any inconsistencies in the input sequence are inherited to the outputs. Since we want to synthesize well-aligned images, and our models don't do so implicitly, we have to handle alignment as a preprocessing step.

For \Valley{}, we perform automatic alignment using LoFTR \cite{Sun2021LoFTR} by matching all frames to an anchor frame, and fitting an affine transform to the detected keypoints. We discard keypoints with confidence $p < 0.5$, and only fit an affine if $N \geq 30$ keypoints are detected, otherwise falling back to the closest valid preceding alignment in the sequence.

The LoFTR-based alignment is quite sensitive to scene content, anchor frame, and choice of parameters. As such, for the other datasets, we instead perform manual alignment: the time-lapse sequence is split at each large discontinuity in alignment, generating several sub-sequences which are internally more consistent. Then, a representative frame is chosen from each sequence, and it is aligned to a global anchor by hand-picking the same three points in both images, and fitting a partial affine transformation with 4 degrees of freedom (translation, rotation, and uniform scale) to the point sets.

\FloatBarrier   
\end{document}

%% file: macros.tex
\newcommand{\bn}{\boldsymbol{n}}
\newcommand{\bw}{\boldsymbol{w}}

\newcommand{\bz}{\boldsymbol{z}}

\newcounter{commentH}
\newcommand{\randomcolor}{%
  \definecolor{randomcolor}{HSB}
   {
    \the\value{commentH},
    256,
    160
   }%
  \color{randomcolor}%
 }
\newcommand{\newrandomcolor}{%
  \addtocounter{commentH}{75}%
  \ifnum\value{commentH}>255%
      \addtocounter{commentH}{-256}%
  \fi%
}

\newcounter{commentcounter}
\newif\ifinlinecomments

\inlinecommentstrue

\ifinlinecomments

\else

\fi

\definecolor{olive}{rgb}{0.5, 0.5, 0.0}
\definecolor{maroon}{rgb}{0.69, 0.19, 0.38}
\definecolor{celestialblue}{rgb}{0.29, 0.59, 0.82}
\definecolor{darkgreen}{rgb}{0.0, 0.5, 0.0}
\definecolor{darkred}{rgb}{0.5, 0.0, 0.0}
\definecolor{grey}{rgb}{0.5,0.5,0.5}
\definecolor{darkblue}{rgb}{0.19, 0.19, 0.62}
\definecolor{silver}{rgb}{0.7,0.7,0.7}

\def\clap#1{\hbox to 0pt{\hss #1\hss}}%
\newcommand{\radd}[1]{#1} 
\newcommand{\rdel}[1]{} 

\newcommand{\zz}{{\bf z}}
\newcommand{\ww}{{\bf w}}
\newcommand{\ZZ}{\mathcal{Z}}
\newcommand{\WW}{\mathcal{W}}

\newcommand{\h}{0mm}
\newcommand{\hh}{0mm}
\newcommand{\hhh}{0mm}

\newcommand{\s}{\hphantom{0}}
\newcommand{\sss}{\hphantom{000}}

\newcommand{\Barn}{\textsc{Barn}}

\newcommand{\Frankfurt}{\textsc{Frankfurt}}
\newcommand{\Kuessnacht}{\textsc{Küssnacht}}
\newcommand{\Muotathal}{\textsc{Muotathal}}
\newcommand{\Normandy}{\textsc{Normandy}}

\newcommand{\Teton}{\textsc{Teton}}
\newcommand{\Twomedicine}{\textsc{Two Medicine}}
\newcommand{\Valley}{\textsc{Valley}}
\newcommand{\MPP}{\textsc{Mielipidepalsta}} 

\makeatletter
\newcommand\cdashfill{\leavevmode\cleaders\hb@xt@.74em{\hss$-$\hss}\hfill\kern\z@}
\newcommand\cdotfill{\leavevmode\cleaders\hb@xt@.44em{\hss$\cdot$\hss}\hfill\kern\z@}
\makeatother

%% file: figures.tex
\newcommand{\hdrbox}[2]{\raisebox{0mm}[\height][.2ex]{\makebox[#1]{\footnotesize$\longleftarrow\cdotfill$ #2 $\cdotfill\longrightarrow$}}}

\newcommand{\figTeaser}[1]{
\begin{teaserfigure}
    \includegraphics[trim=30 60 30 61,clip,width=\textwidth]{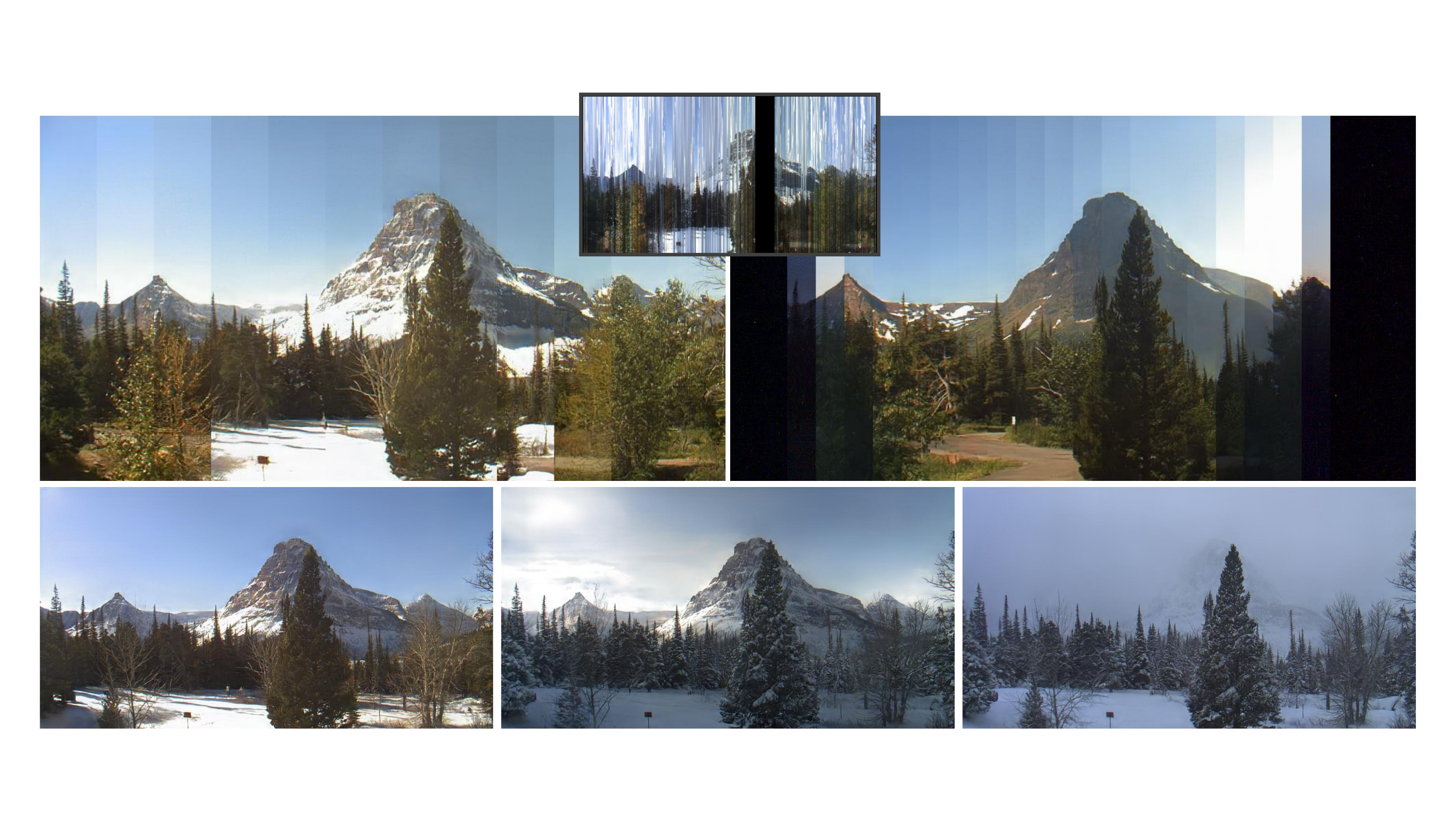}
    \raisebox{86mm}[0mm][0mm]{\makebox[0.45\linewidth]{Synthesized year cycle}}\hfill%
    \raisebox{86mm}[0mm][0mm]{\makebox[0.45\linewidth]{Synthesized day cycle}}\vspace{-2.3\baselineskip}\\
    \caption{
    Our goal is to take a captured time-lapse sequence (top-middle, one year in \Twomedicine{} shown) and learn a generative model using it. Such real datasets typically contains a lot of flickering due to weather changes and missing frames. We can use the generative model to synthesize various cleaner versions of the sequence, such as a year cycle or a day cycle under coherent weather, visualized on the top row by choosing image columns from different parts of the synthetic sequences. On the bottom row, we show three realizations of a frame captured on 17.12.2015 12:30, synthesized using three different weathers that were learned from the input sequence.
    }\vspace{0.09\baselineskip}
    \Description{A collage of four images showing a lake scene as visualized using the dataset frames, and our trained models.}
    \label{#1}
\end{teaserfigure}
}

\newcommand{\figTimeOfDay}[1]{
\renewcommand{\h}{0.240\textwidth}
\begin{figure*}
    \centering

    \rotatebox[origin=l]{90}{\makebox[0mm][l]{\normalsize{\phantom{\Barn}}}}\hfill%
    \hdrbox{\h}{18 hours}\hfill\hfill\hfill\hdrbox{\h}{18 hours}\hfill\hdrbox{\h}{18 hours}\hfill\hdrbox{\h}{18 hours}\vspace{-1.7mm}\\
    \rotatebox[origin=l]{90}{\makebox[0mm][l]{\normalsize{\phantom{\Barn}}}}\hfill%
    \makebox[\h]{\tiny Dawn \hspace*{10mm} Noon \hspace*{10mm} Dusk}\hfill\hfill\hfill%
    \makebox[\h]{\tiny Dawn \hspace*{10mm} Noon \hspace*{10mm} Dusk}\hfill%
    \makebox[\h]{\tiny Dawn \hspace*{10mm} Noon \hspace*{10mm} Dusk}\hfill%
    \makebox[\h]{\tiny Dawn \hspace*{10mm} Noon \hspace*{10mm} Dusk}\\
    
    \rotatebox[origin=l]{90}{\makebox[0mm][l]{\hspace*{0.008\linewidth}\Frankfurt{}\hspace{1mm}\footnotesize{6.4.2015}}}\hfill%
    \includegraphics[trim=0 20 0 60,clip,width=\h]{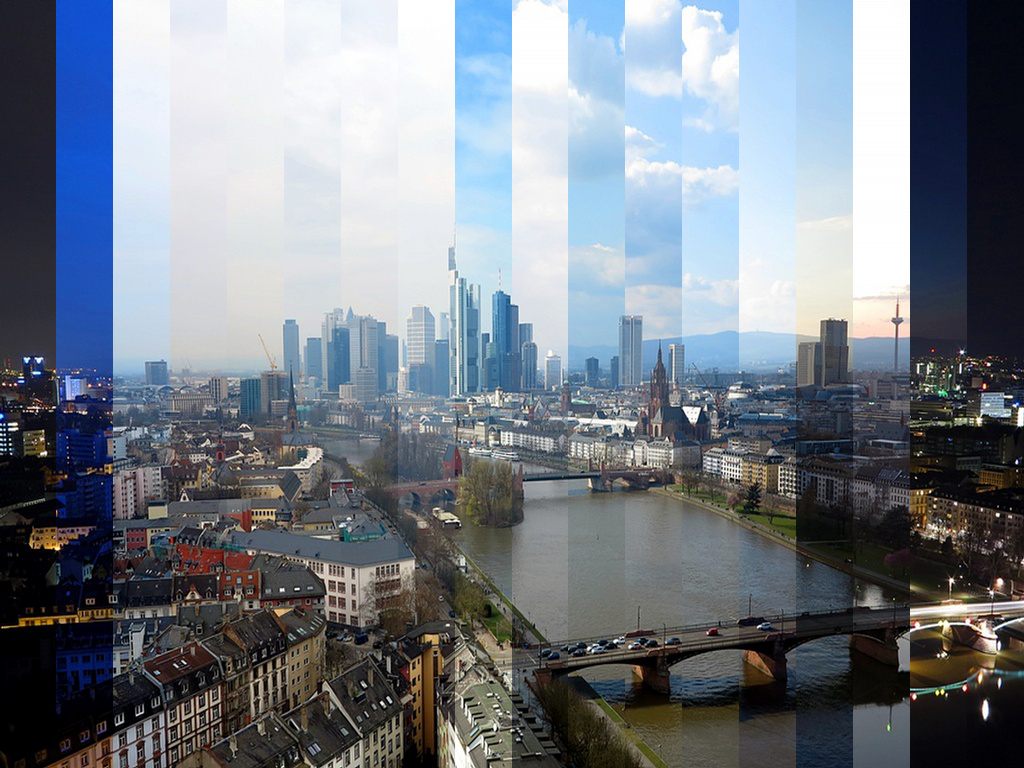}\hfill\hfill\hfill%
    \includegraphics[trim=0 20 0 60,clip,width=\h]{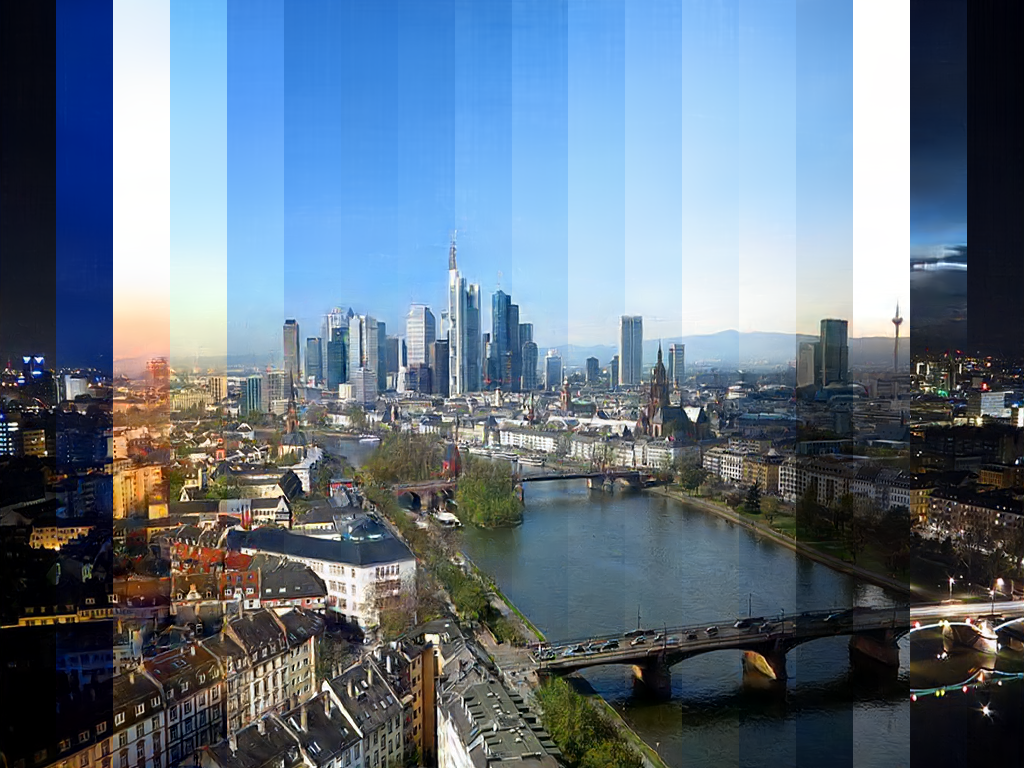}\hfill%
    \includegraphics[trim=0 20 0 60,clip,width=\h]{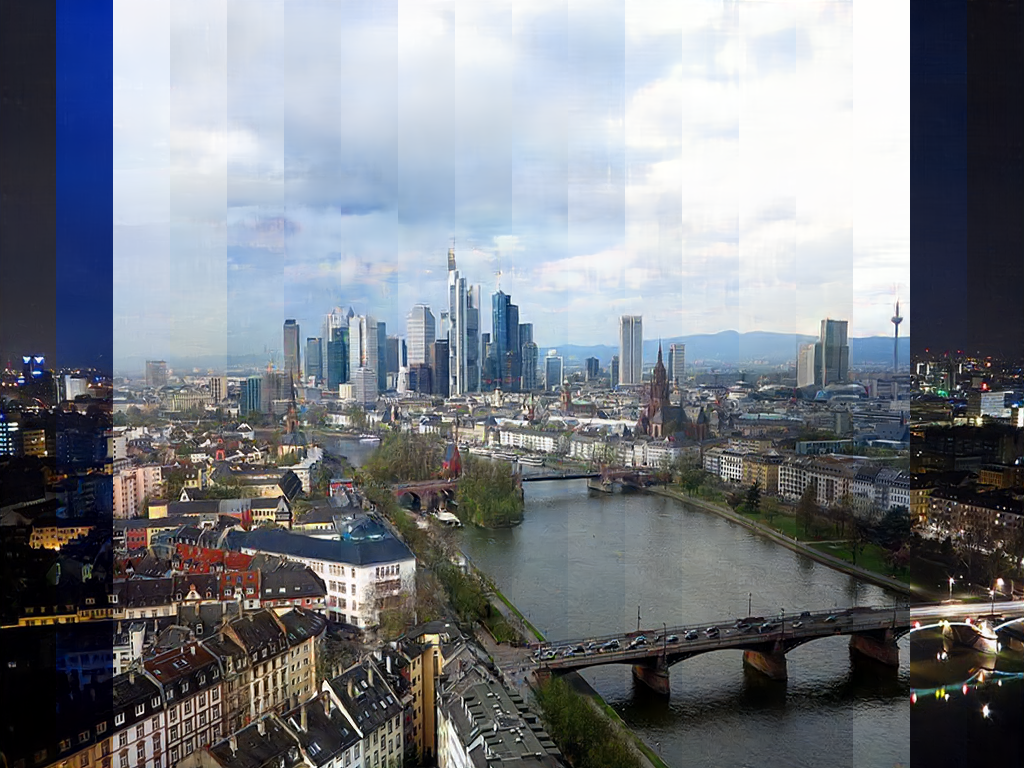}\hfill%
    \includegraphics[trim=0 20 0 60,clip,width=\h]{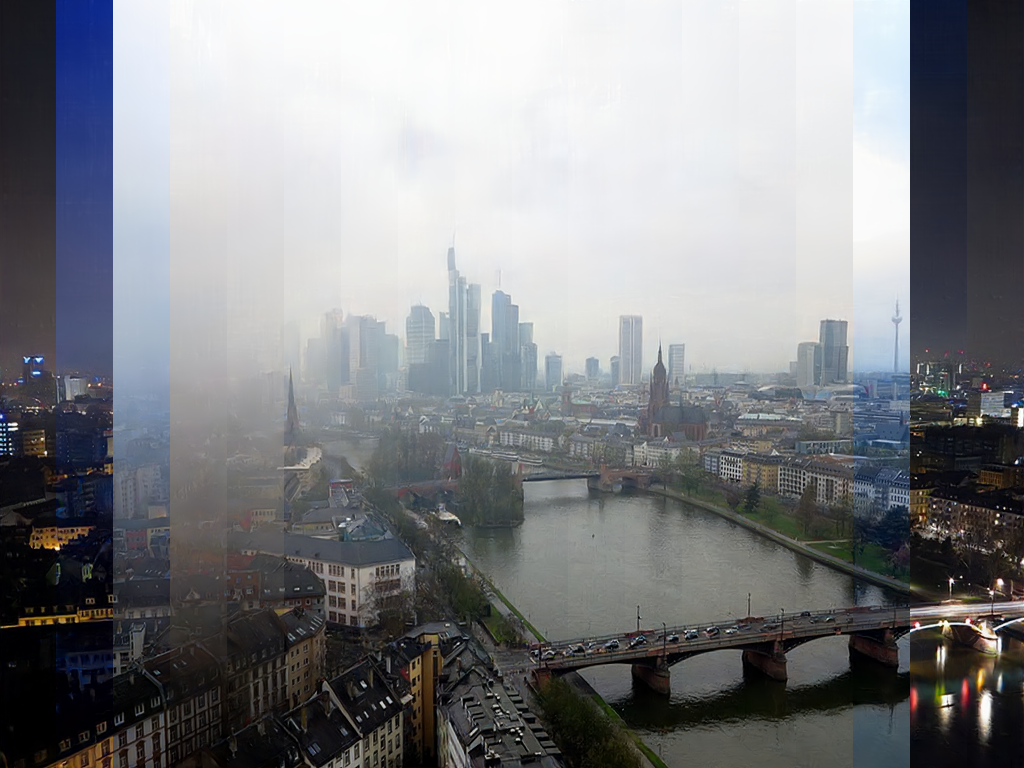}
    
    \rotatebox[origin=l]{90}{\makebox[0mm][l]{\hspace*{0.001\linewidth}\Kuessnacht{}\hspace{1mm} \footnotesize{19.2.2013}}}\hfill%
    \includegraphics[trim=0 0 0 10,clip,width=\h]{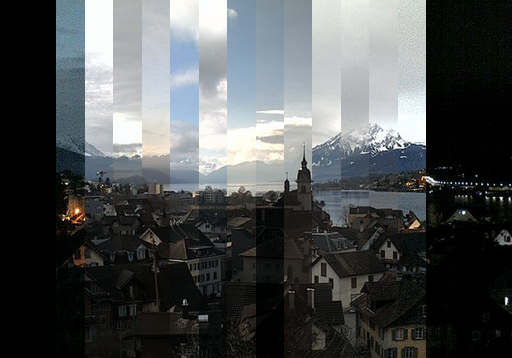}\hfill\hfill\hfill%
    \includegraphics[trim=0 0 0 10,clip,width=\h]{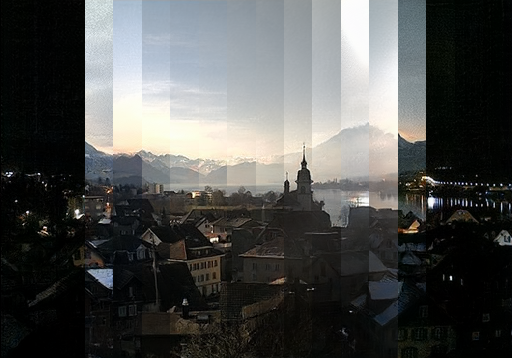}\hfill%
    \includegraphics[trim=0 0 0 10,clip,width=\h]{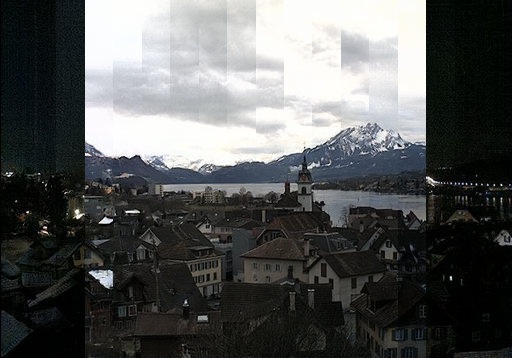}\hfill%
    \includegraphics[trim=0 0 0 10,clip,width=\h]{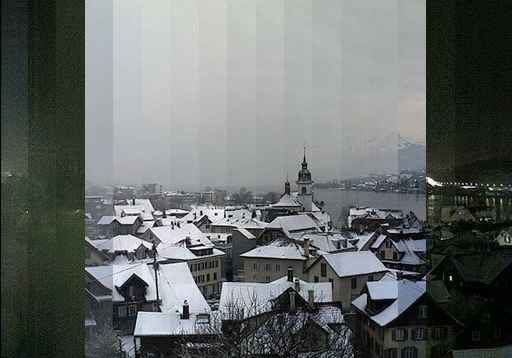}
    
    \rotatebox[origin=l]{90}{\makebox[0mm][l]{\hspace*{0.02\linewidth}\Teton{}\hspace{1mm}\footnotesize{13.10.2012}}}\hfill%
    \includegraphics[width=\h]{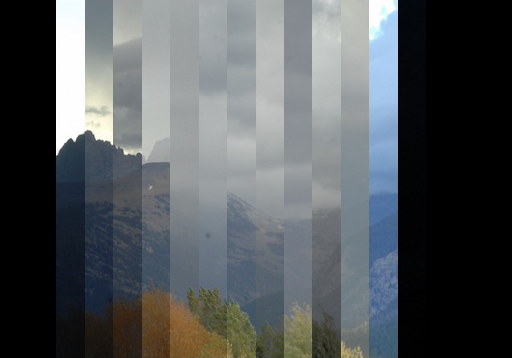}\hfill\hfill\hfill%
    \includegraphics[width=\h]{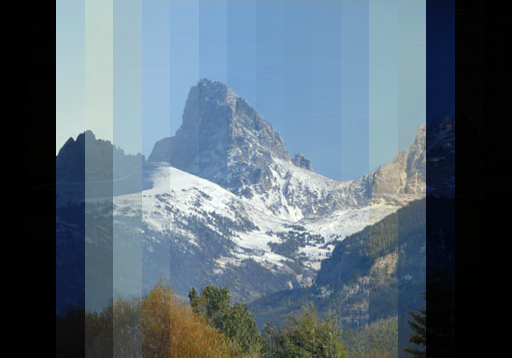}\hfill%
    \includegraphics[width=\h]{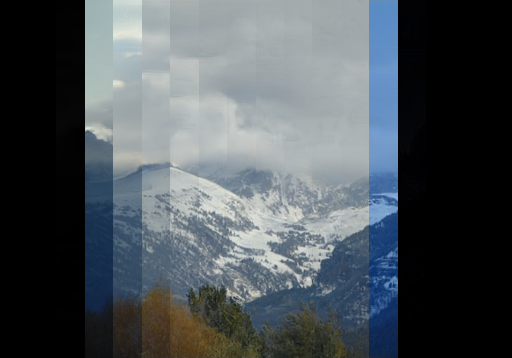}\hfill%
    \includegraphics[width=\h]{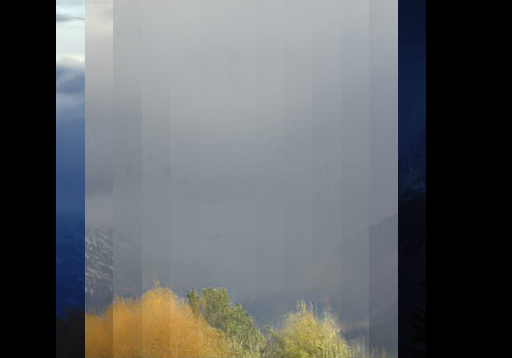}\vspace{0.7mm}\\
    
    \rotatebox[origin=l]{90}{\makebox[0mm][l]{\hspace*{0.05\linewidth}\normalsize{\phantom{\Barn}}}}\hfill%
    \hdrbox{\h}{12 months}\hfill\hfill\hfill\hdrbox{\h}{12 months}\hfill\hdrbox{\h}{12 months}\hfill\hdrbox{\h}{12 months}\vspace{-1.7mm}\\
    \rotatebox[origin=l]{90}{\makebox[0mm][l]{\normalsize{\phantom{\Barn}}}}\hfill%
    \makebox[\h]{\tiny Summer \hspace*{2.8mm} Autumn \hspace*{2.8mm} Winter \hspace*{2.8mm} Spring \hspace*{2.8mm}  Summer}\hfill\hfill\hfill%
    \makebox[\h]{\tiny Summer \hspace*{2.8mm} Autumn \hspace*{2.8mm} Winter \hspace*{2.8mm} Spring \hspace*{2.8mm}  Summer}\hfill%
    \makebox[\h]{\tiny Summer \hspace*{2.8mm} Autumn \hspace*{2.8mm} Winter \hspace*{2.8mm} Spring \hspace*{2.8mm}  Summer}\hfill%
    \makebox[\h]{\tiny Summer \hspace*{2.8mm} Autumn \hspace*{2.8mm} Winter \hspace*{2.8mm} Spring \hspace*{2.8mm}  Summer}\\
   
    \rotatebox[origin=l]{90}{\makebox[0mm][l]{\hspace*{0.03\linewidth}\Valley\hspace{1mm}\footnotesize{2016}}}\hfill%
    \includegraphics[trim=0 0 10 150,clip,width=\h]{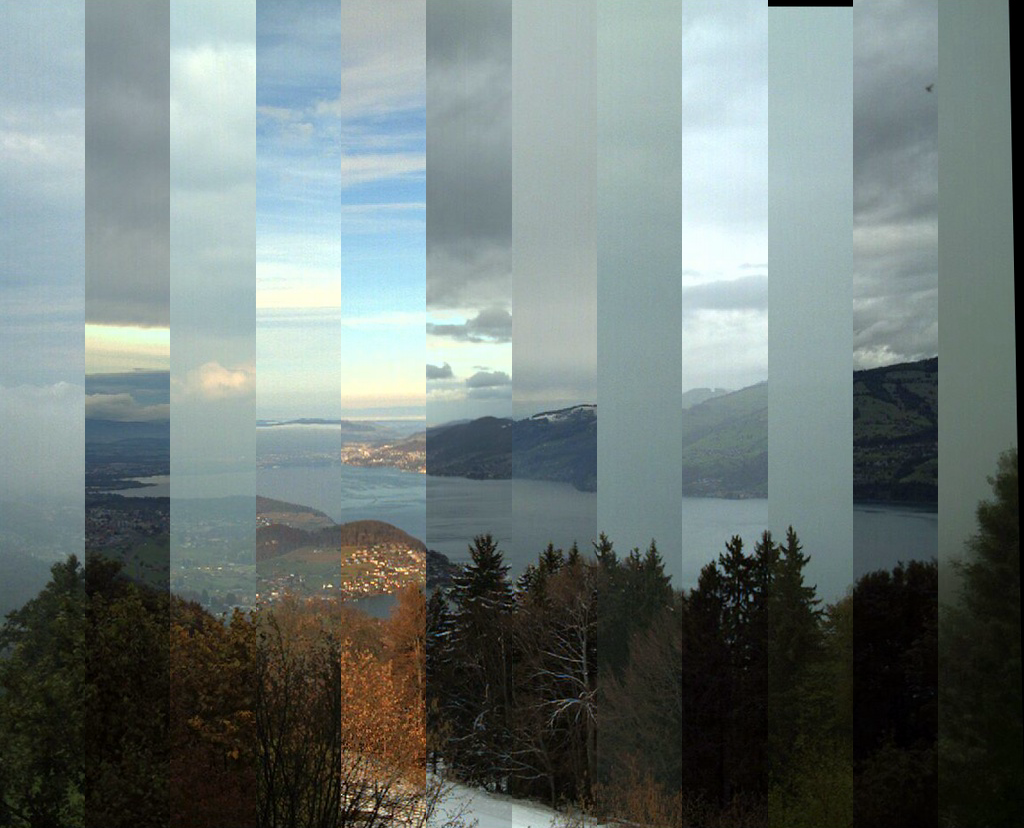}\hfill\hfill\hfill%
    \includegraphics[trim=0 0 10 150,clip,width=\h]{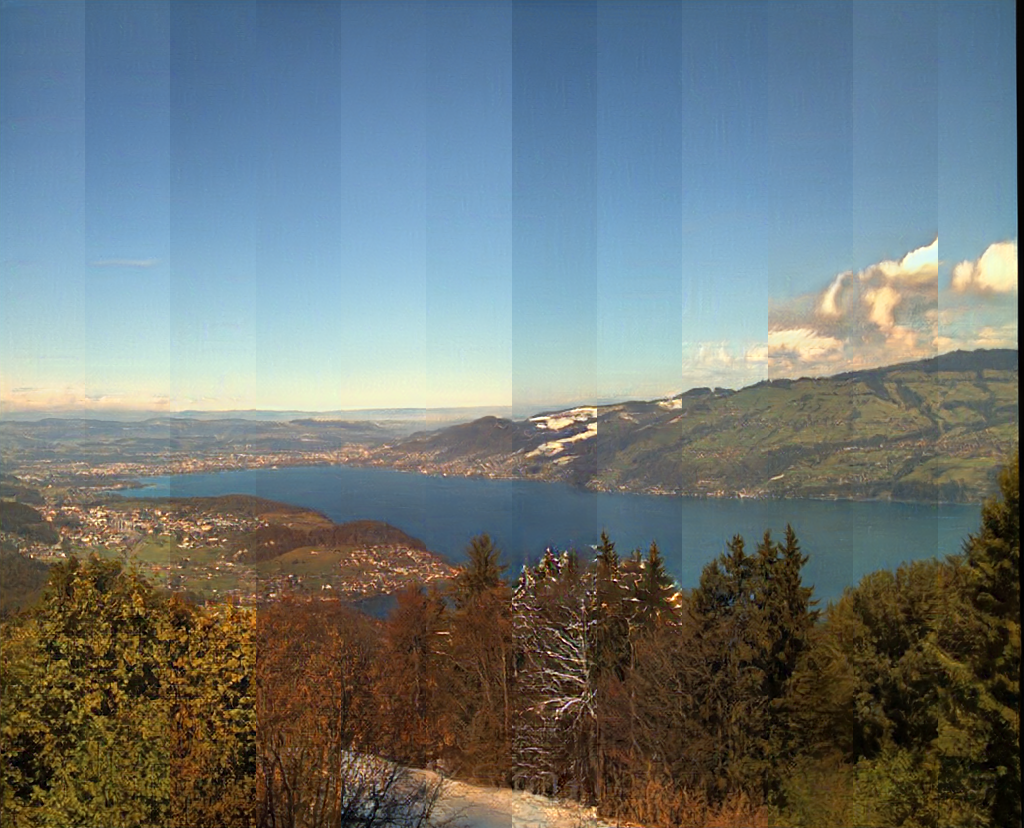}\hfill%
    \includegraphics[trim=0 0 10 150,clip,width=\h]{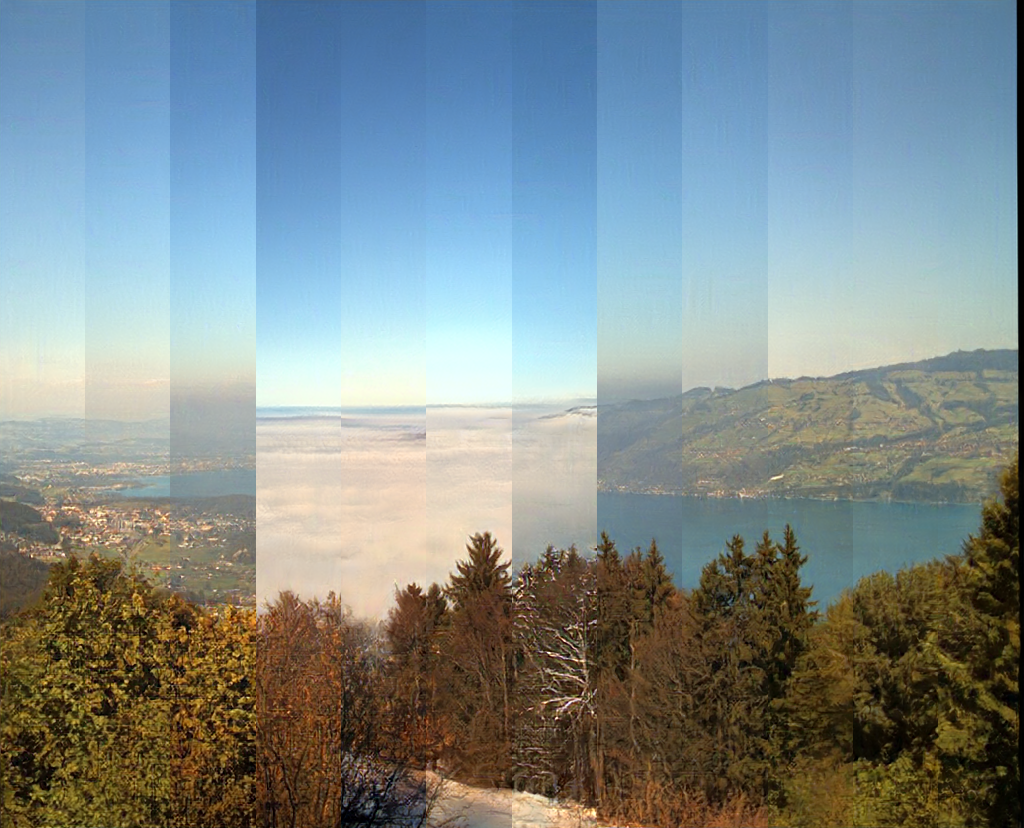}\hfill%
    \includegraphics[trim=0 0 10 150,clip,width=\h]{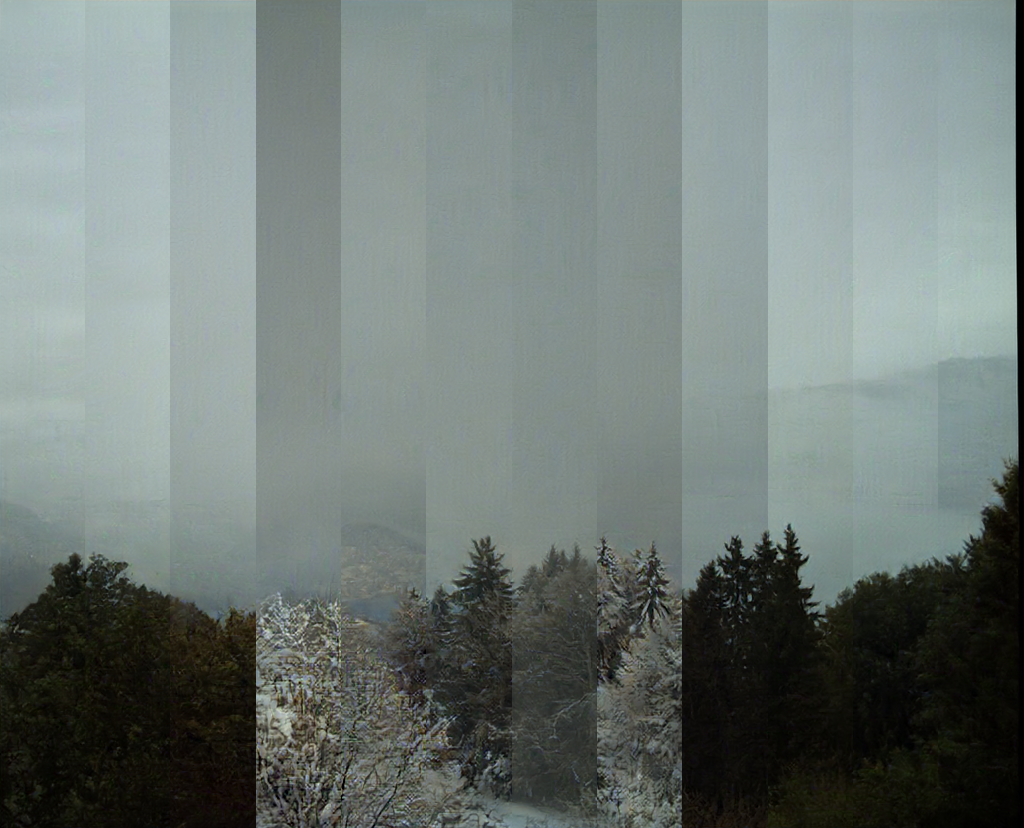}
    
    \rotatebox[origin=l]{90}{\makebox[0mm][l]{\hspace*{-0.002\linewidth}\Twomedicine\hspace{1mm}\footnotesize{2016}}}\hfill%
    \includegraphics[trim=73 0 73 0,clip,width=\h]{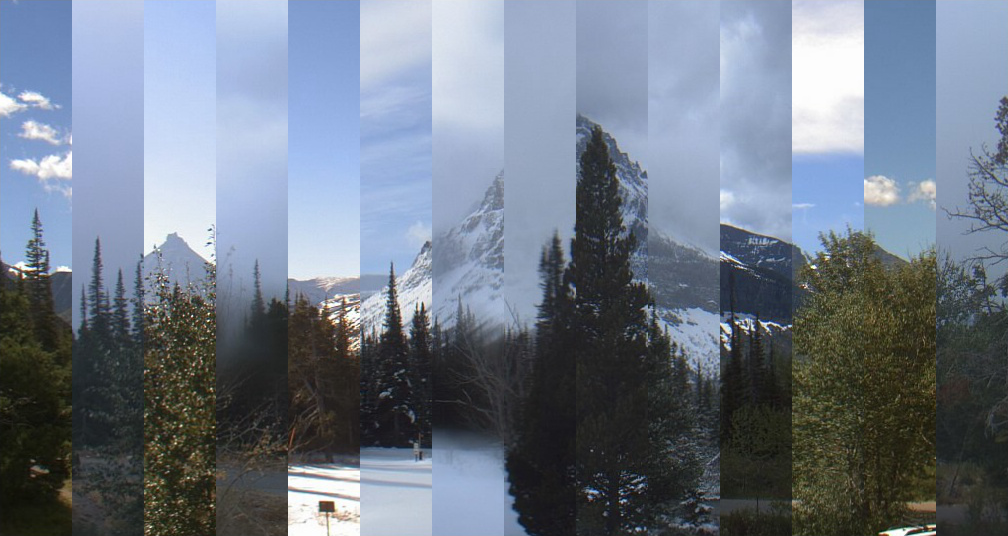}\hfill\hfill\hfill%
    \includegraphics[trim=73 0 73 0,clip,width=\h]{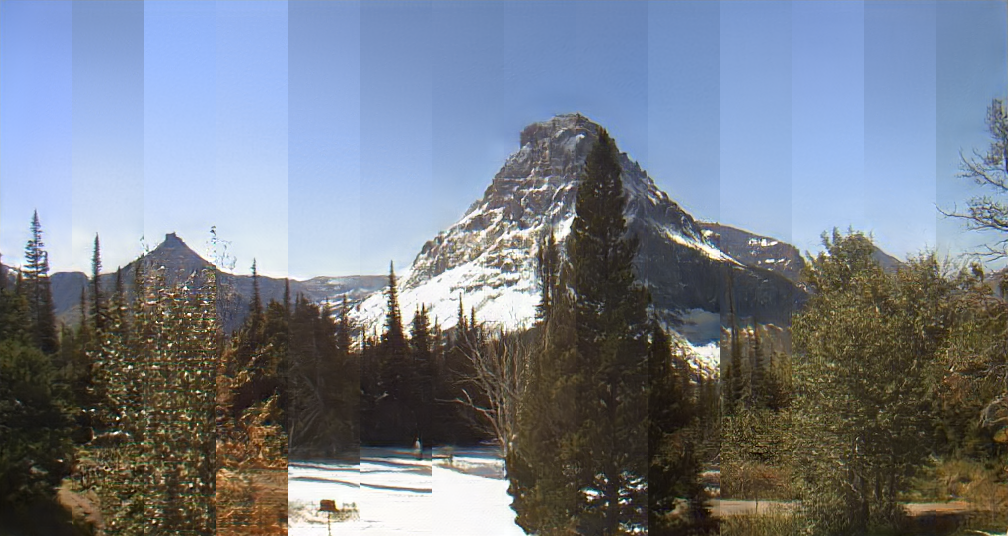}\hfill%
    \includegraphics[trim=73 0 73 0,clip,width=\h]{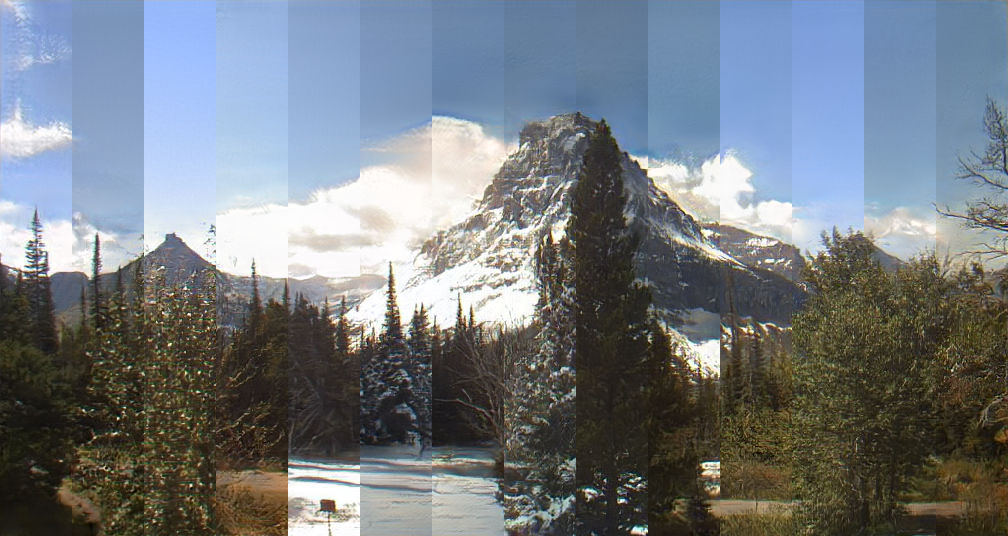}\hfill%
    \includegraphics[trim=73 0 73 0,clip,width=\h]{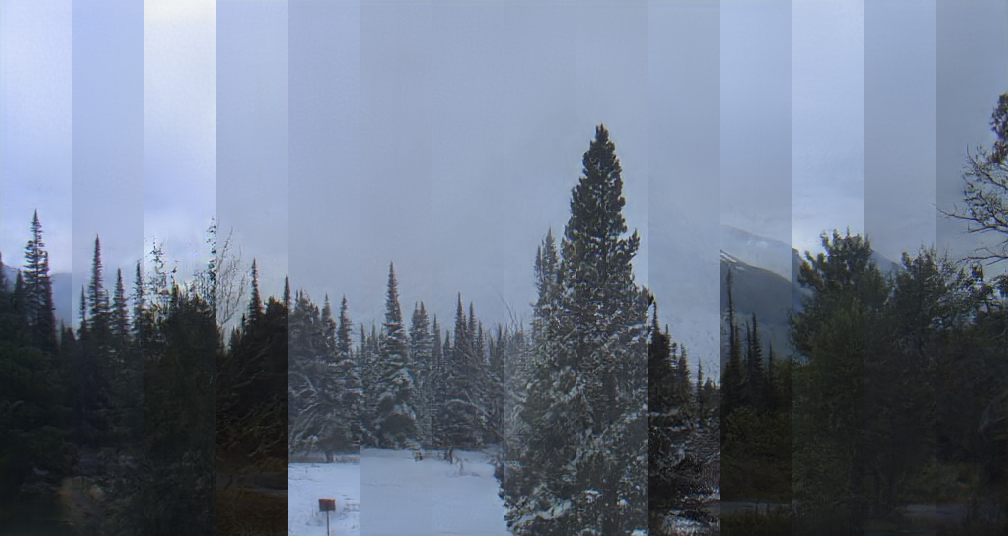}
    
    \rotatebox[origin=l]{90}{\makebox[0mm][l]{\hspace*{0.02\linewidth}\Muotathal{}\hspace{1mm}\footnotesize{2013}}}\hfill%
    \includegraphics[trim=0 65 0 0,clip,width=\h]{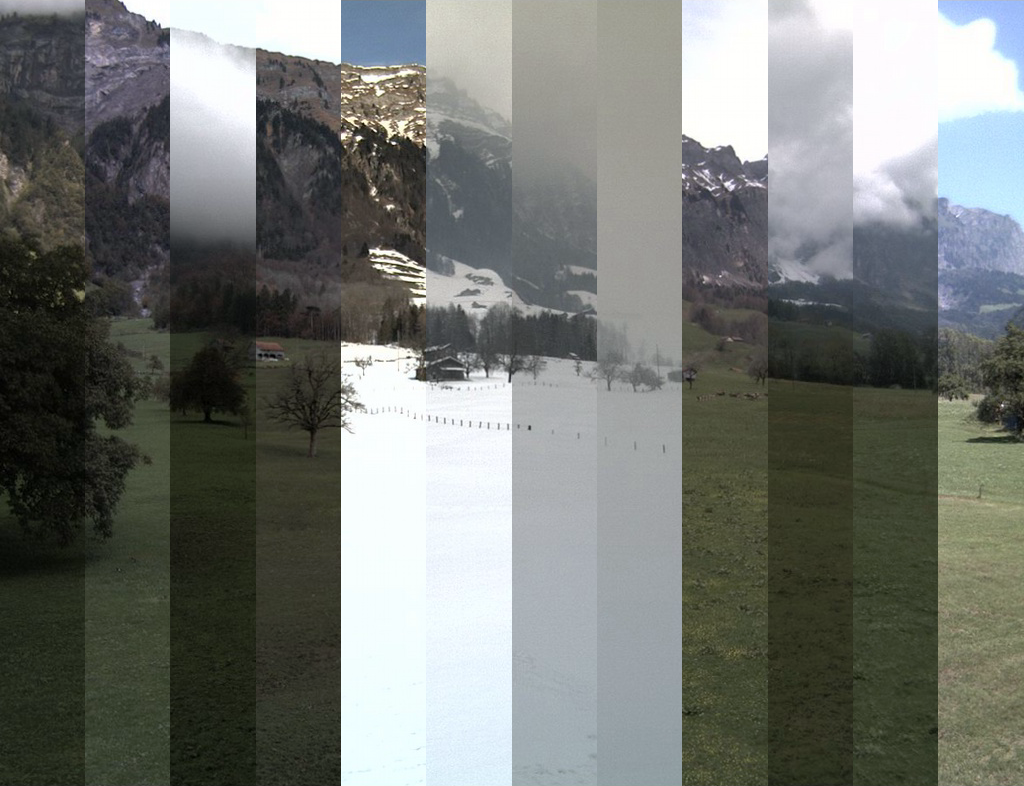}\hfill\hfill\hfill%
    \includegraphics[trim=0 65 0 0,clip,width=\h]{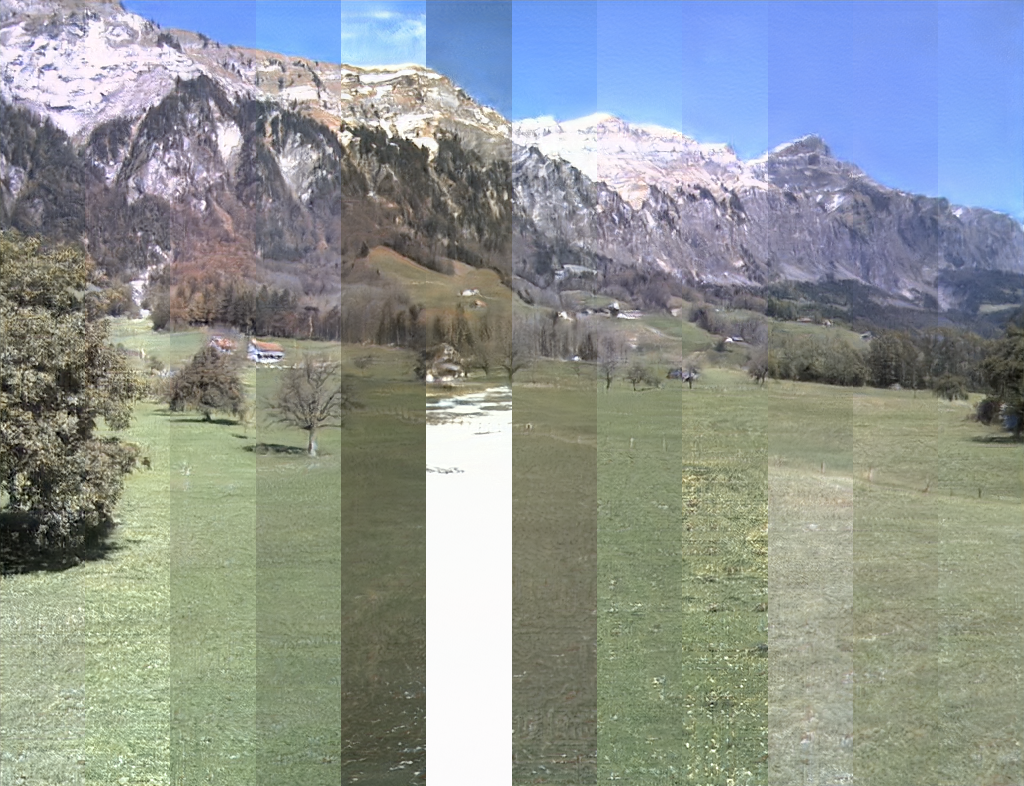}\hfill%
    \includegraphics[trim=0 65 0 0,clip,width=\h]{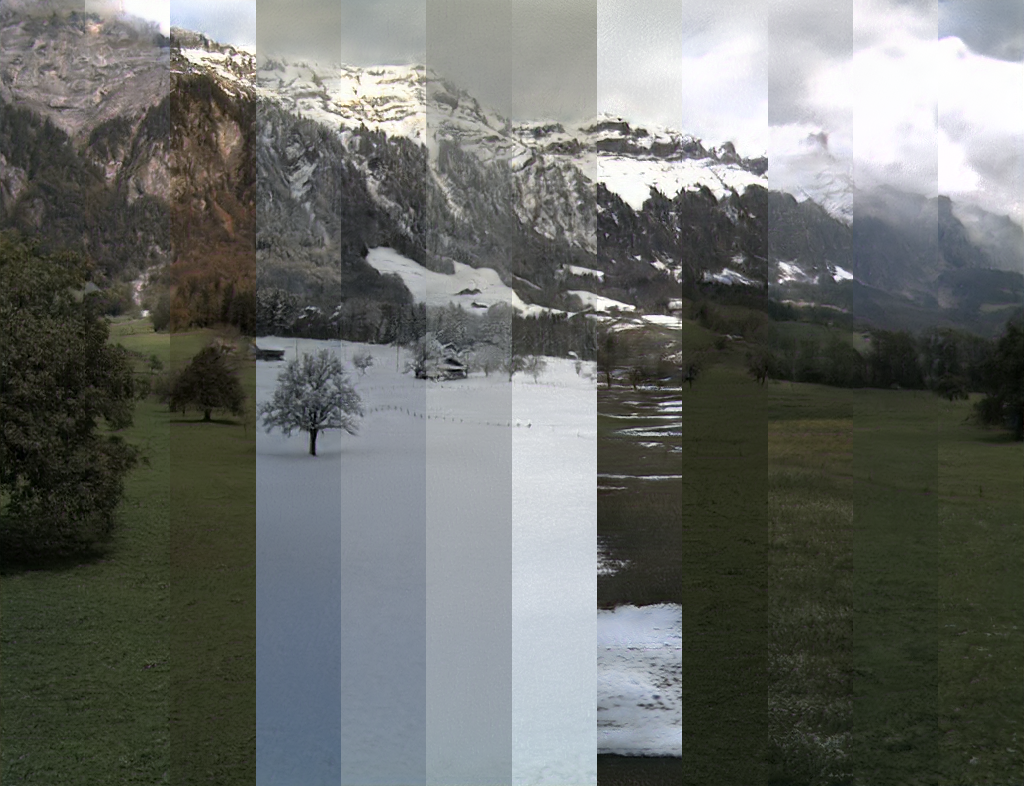}\hfill%
    \includegraphics[trim=0 65 0 0,clip,width=\h]{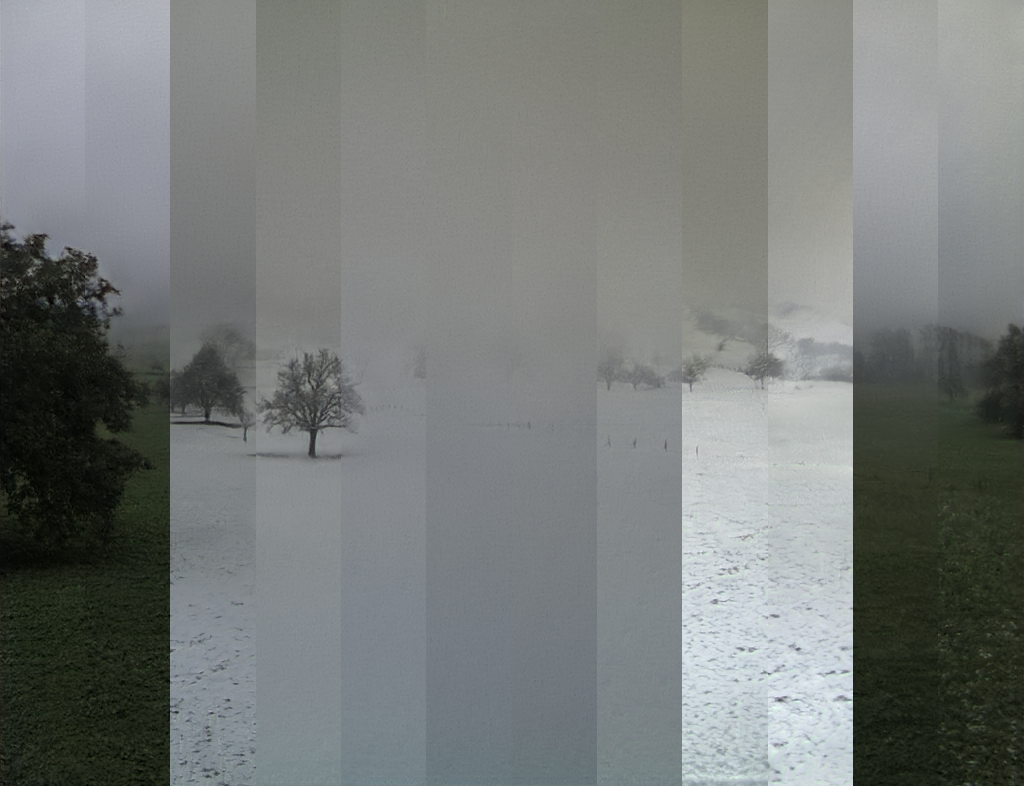}\vspace{-0.5mm}\\

    \makebox[\h]{Input data}
    \makebox[\h]{Ours, clear skies}
    \makebox[\h]{Ours, clouds}
    \makebox[\h]{Ours, fog/snow}\vspace{-0.6mm}\\
    \caption{In these six datasets, we change only the time-of-day (top 3 rows) or time-of-year (bottom 3 rows) conditioning signal. 
    The input data shows the time-lapse image for that particular day or year, respectively.
    While the inputs have reasonably constant weather over a day, the same is not at all true for the whole year.
    When we sweep the time-of-day signal, the synthesized time-lapse images show a clear day cycle, as desired. Similarly, the time-of-year causes the time-lapse image to cycle through the seasons (note that winter is in the middle). We furthermore visualize our results using 3 different latent codes, chosen separately for each dataset to demonstrate approximately clear sky, some clouds, and a foggy or snowy appearance. 
    For the top part, we show 18 of the 24 hours to crop out some of the uneventful nighttime. }
    \label{#1}
\end{figure*}
}

\newcommand{\figTide}[1]{
\renewcommand{\h}{0.121\linewidth}
\begin{figure*}
    \centering
    \rotatebox{90}{\parbox[c][3mm][c]{\h}{\centering Training data}}
    \includegraphics[height=\h]{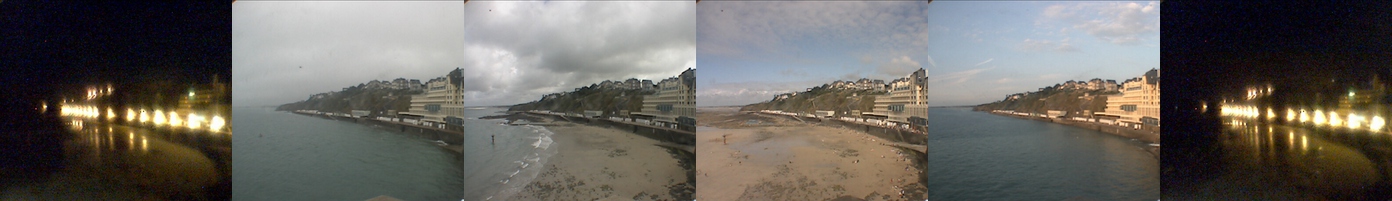}\hfill\\
    \rotatebox{90}{\parbox[c][3mm][c]{\h}{\centering Ours}}
    \includegraphics[height=\h]{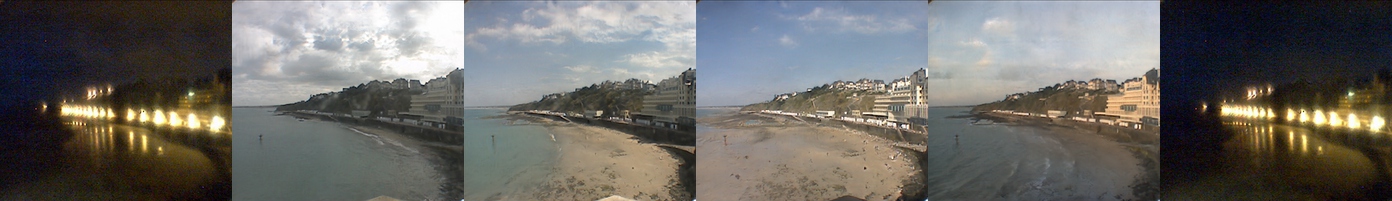}\hfill\\
    \caption{In \Normandy{}, time-of-day controls also the tidal changes.}
    \label{#1}
\end{figure*}
}

\newcommand{\figDayLength}[1]{
\renewcommand{\h}{0.48\columnwidth}
\begin{figure}
    \rotatebox[origin=l]{90}{\makebox[0mm][l]{}}\hfill%
    \hdrbox{\h}{24 hours}\hfill%
    \hdrbox{\h}{24 hours}\\
    \rotatebox[origin=l]{90}{\makebox[0mm][l]{\hdrbox{0.52\columnwidth}{1 year}}}\hfill%
    \includegraphics[trim=100 0 100 0,clip,width=\h]{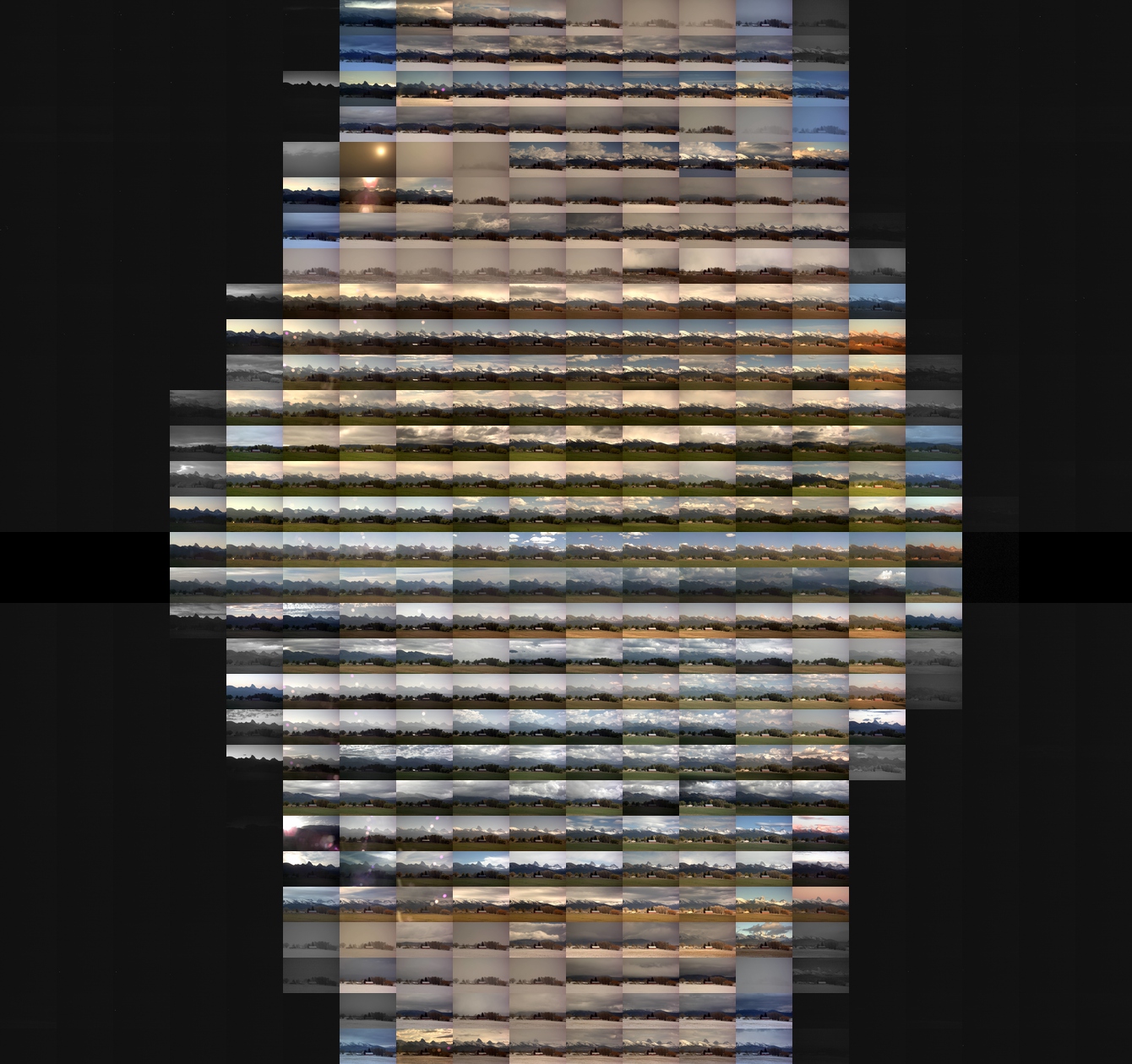}\hfill%
    \includegraphics[trim=100 0 100 0,clip,width=\h]{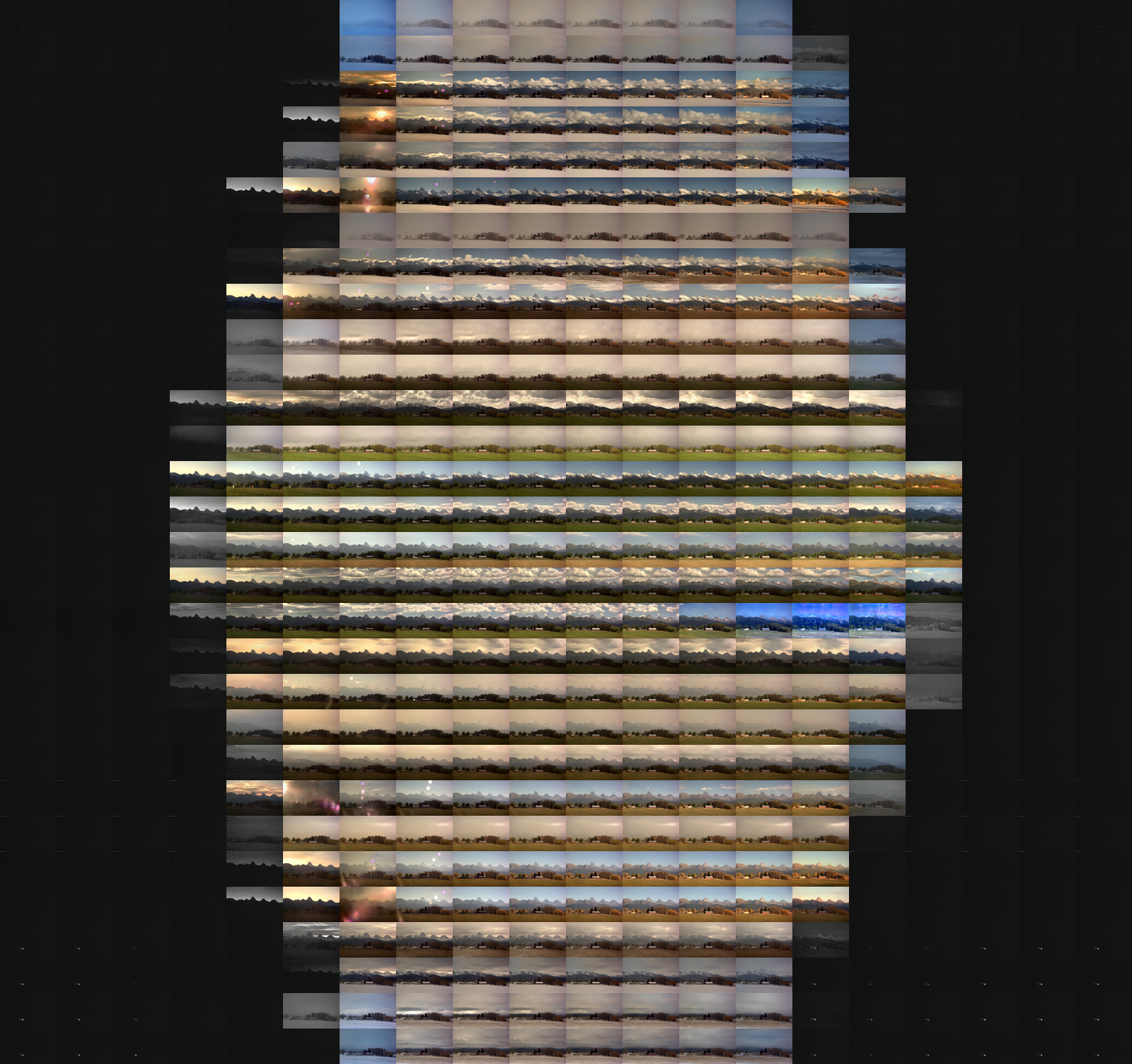}
    \rotatebox[origin=l]{90}{\makebox[0mm][l]{}}\hfill%
    \makebox[\h]{Training data}\hfill%
    \makebox[\h]{Ours}
    \caption{Our model reproduces seasonal changes in day length, shown above by varying the 24h cycle (horizontally) and the year cycle (vertically). Our result is computed using a different random latent code $\bz$ per row.}
    \label{#1}
\end{figure}
}

\newcommand{\figIssues}[1]{
\begin{figure}
    \renewcommand{\h}{0.495\columnwidth}
    \hdrbox{\h}{5 years}\hfill%
    \hdrbox{\h}{12 months}\\
    \includegraphics[trim=150 0 0 120,clip,width=\h]{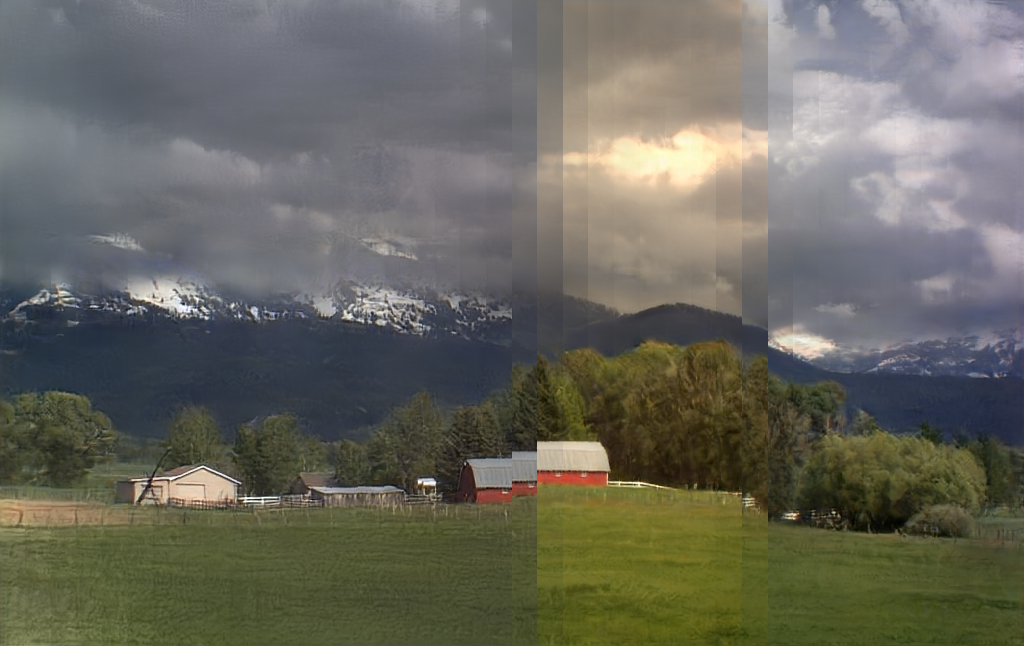}\hfill%
    \includegraphics[trim=0 170 0 0,clip,width=\h]{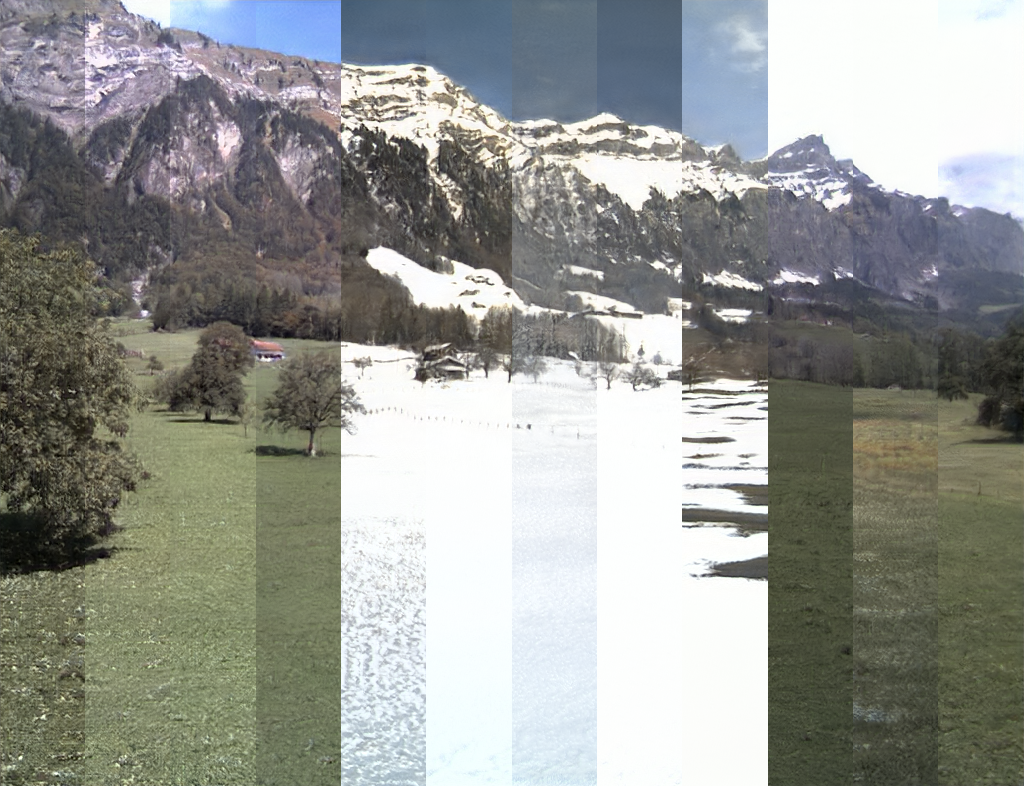}\\
    \makebox[\h]{(a) \Barn{}}\hfill%
    \makebox[\h]{(b) \Muotathal{}}\\
    \caption{Examples of unwanted entanglement. 
    (a) An alignment error that remained after preprocessing was learned by our trend signal, as was the change in color temperature. 
    (b) Automatic brightness adjustment in the camera led to the sky darkening during winters, causing our time-of-year signal to learn this effect.}
    \label{#1}
\end{figure}
}

\newcommand{\figArchitecture}[1]{
\begin{figure}
    \includegraphics[trim=0 130 425 0,clip,width=\columnwidth]{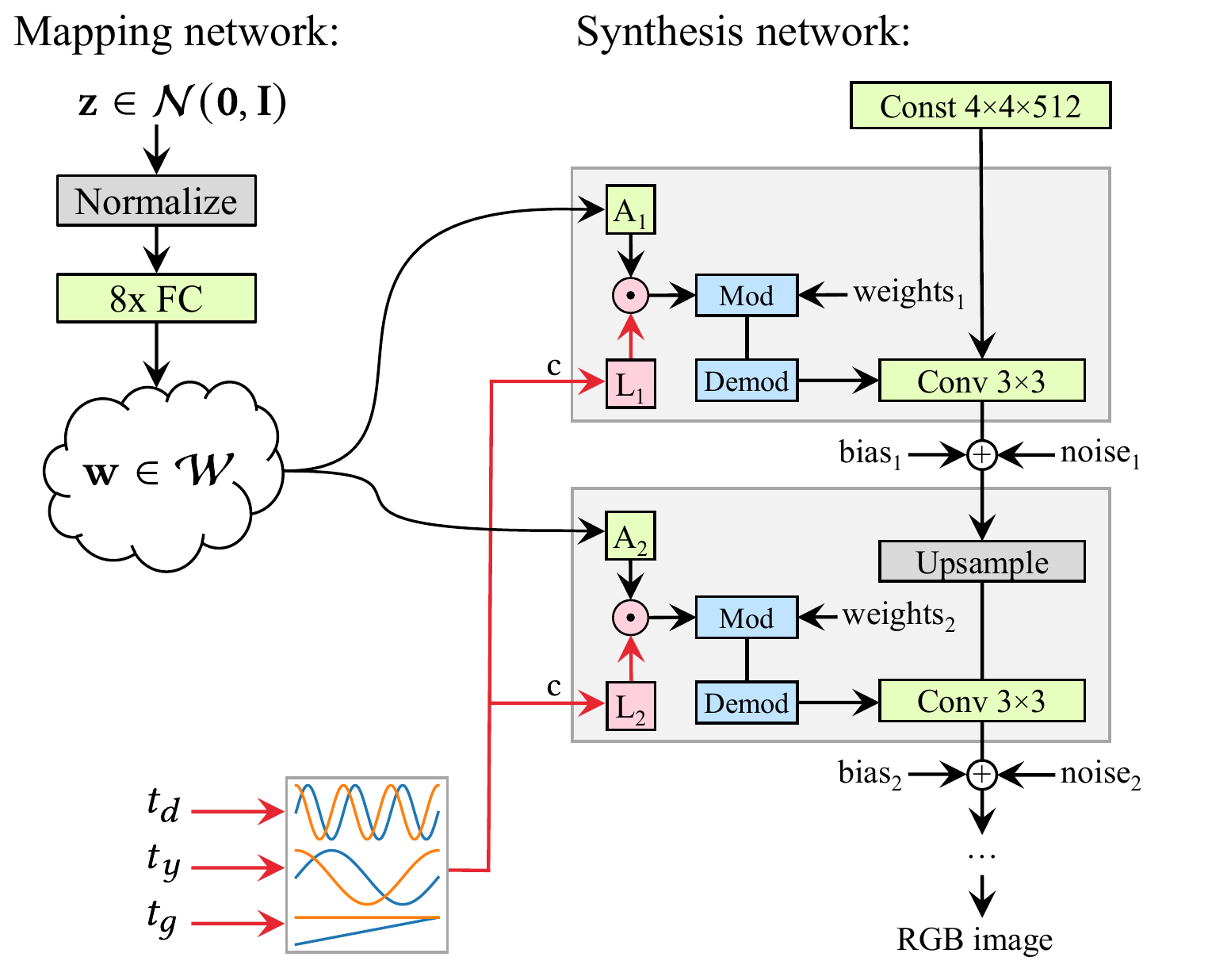}\\
    \caption{StyleGAN2 architecture with our cyclic conditioning mechanism highlighted using red. 
    Our conditioning first transforms a timestamp $(t_d, t_y, t_g)$ into a stack of conditioning signals $\mathbf{c}\radd{=(\mathbf{c}_d, \mathbf{c}_y, \mathbf{c}_g)}$. These are further transformed, at each layer $i$, using a learnable linear transformation $\mathbf{L}_i$ and used to modulate the convolution weights. 
    The StyleGAN2 symbols "FC", $\mathbf{A}$, "Mod", and "Demod" denote a fully connected layer, learned affine transformation, modulation, and demodulation, respectively. 
    }
    \label{#1}
\end{figure}
}

\newcommand{\figLatents}[1]{
\renewcommand{\h}{0.243\textwidth}%
\begin{figure*}
    \centering
    \rotatebox[origin=l]{90}{\makebox[0mm][l]{\hspace{4mm}19.9.2012\hspace{1.5mm}\footnotesize{16:30}}}\hfill%
    \includegraphics[width=\h]{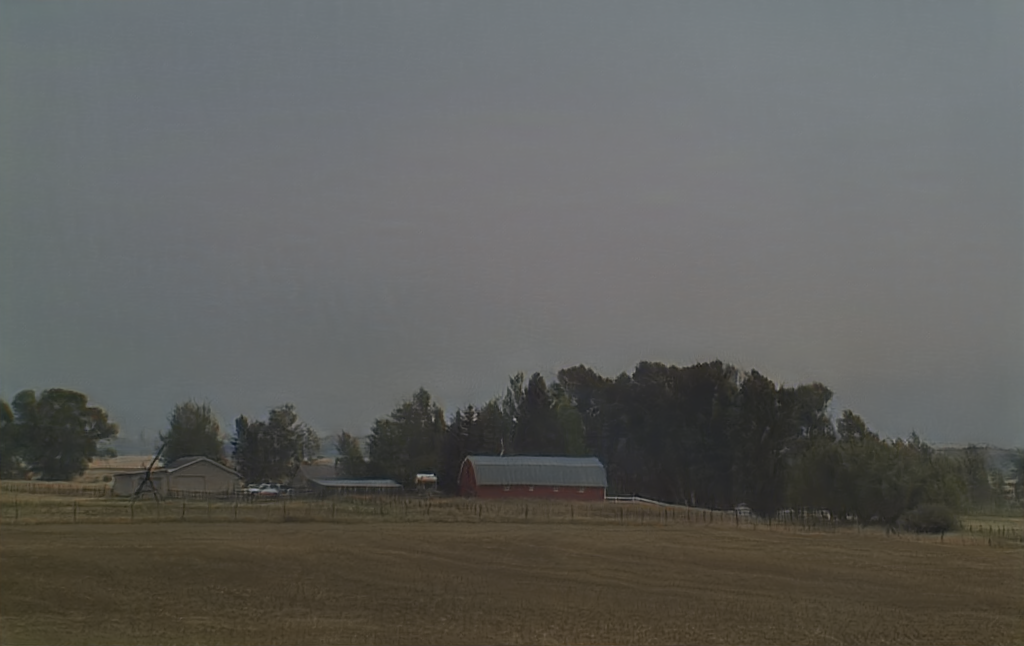}\hfill%
    \includegraphics[width=\h]{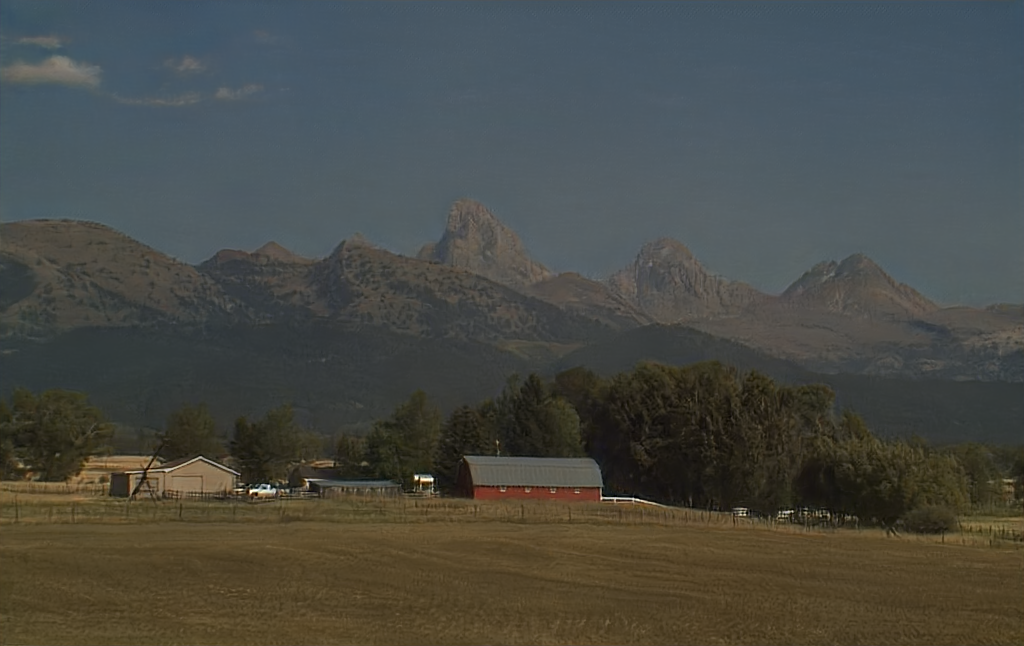}\hfill%
    \includegraphics[width=\h]{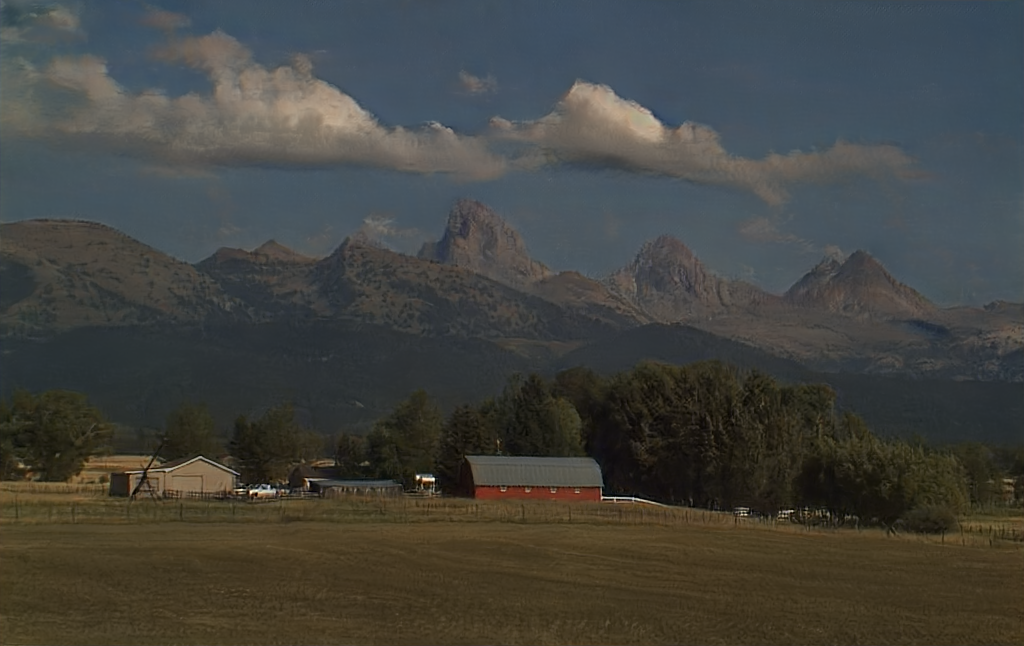}\hfill%
    \includegraphics[width=\h]{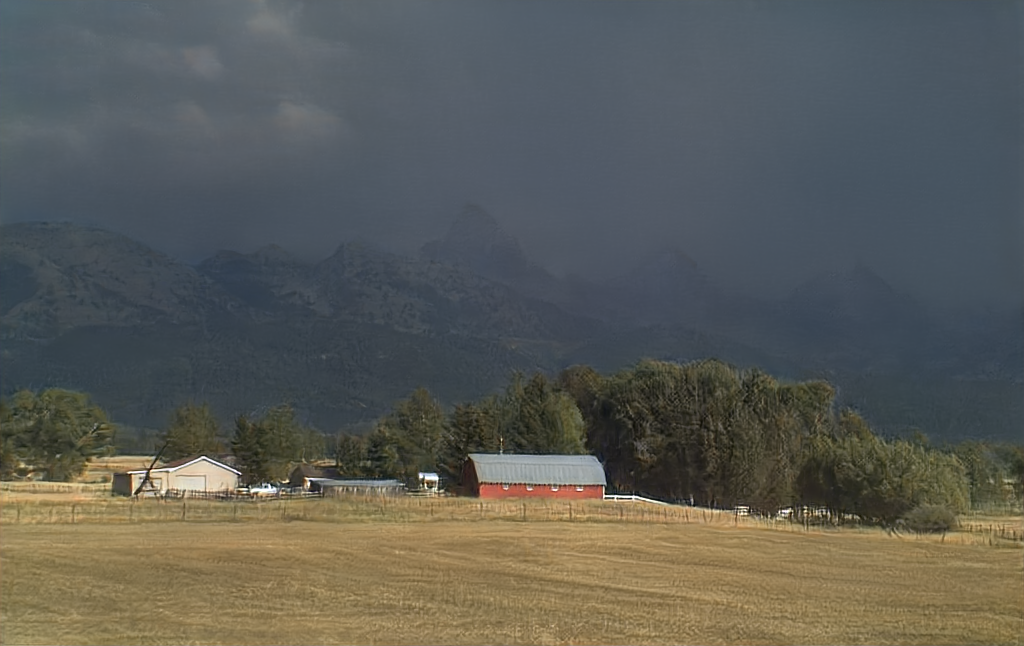}\\%
    \rotatebox[origin=l]{90}{\makebox[0mm][l]{\hspace{4mm}19.8.2015\hspace{1.5mm}\footnotesize{7:30}}}\hfill%
    \includegraphics[width=\h]{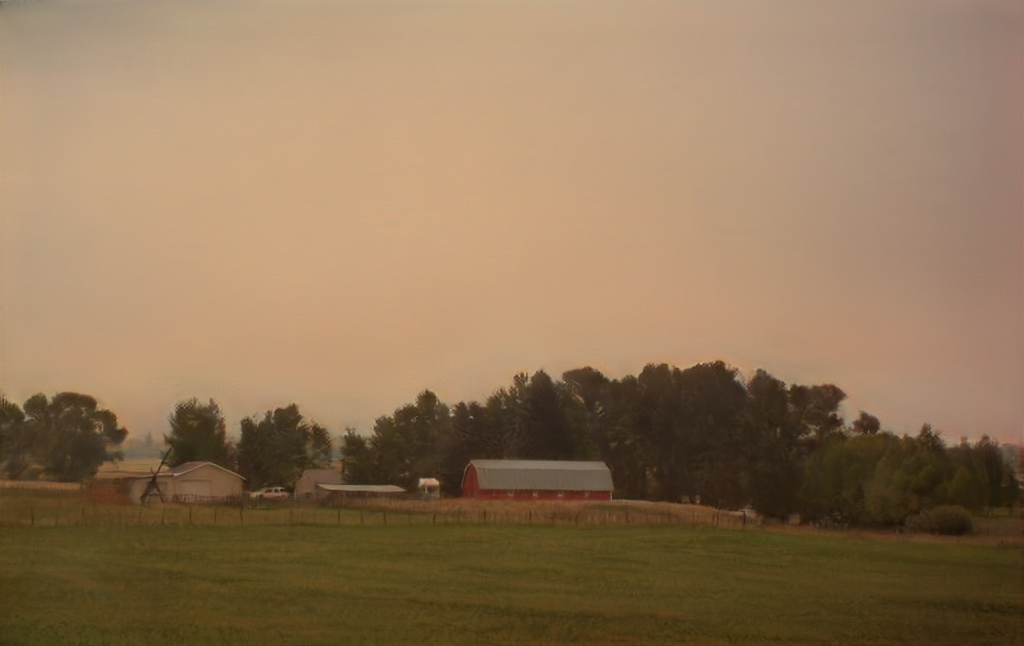}\hfill%
    \includegraphics[width=\h]{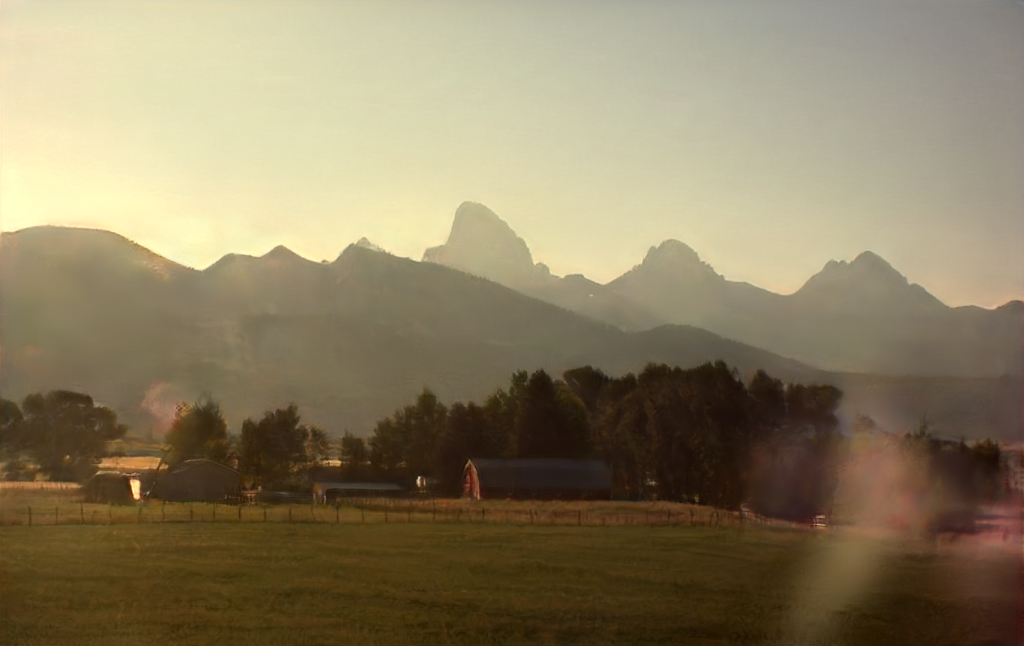}\hfill%
    \includegraphics[width=\h]{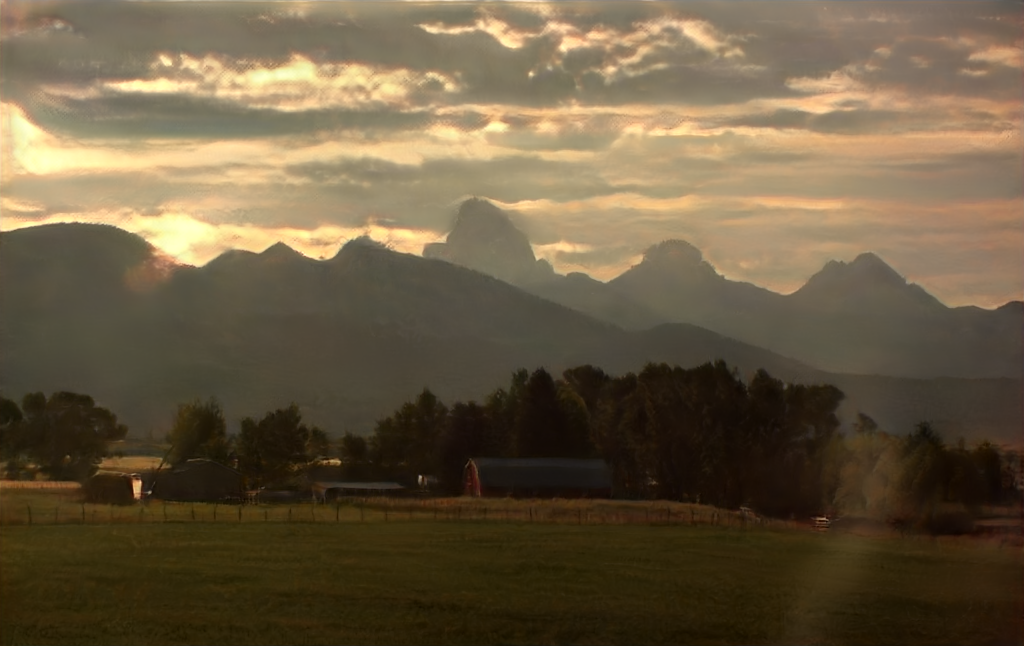}\hfill%
    \includegraphics[width=\h]{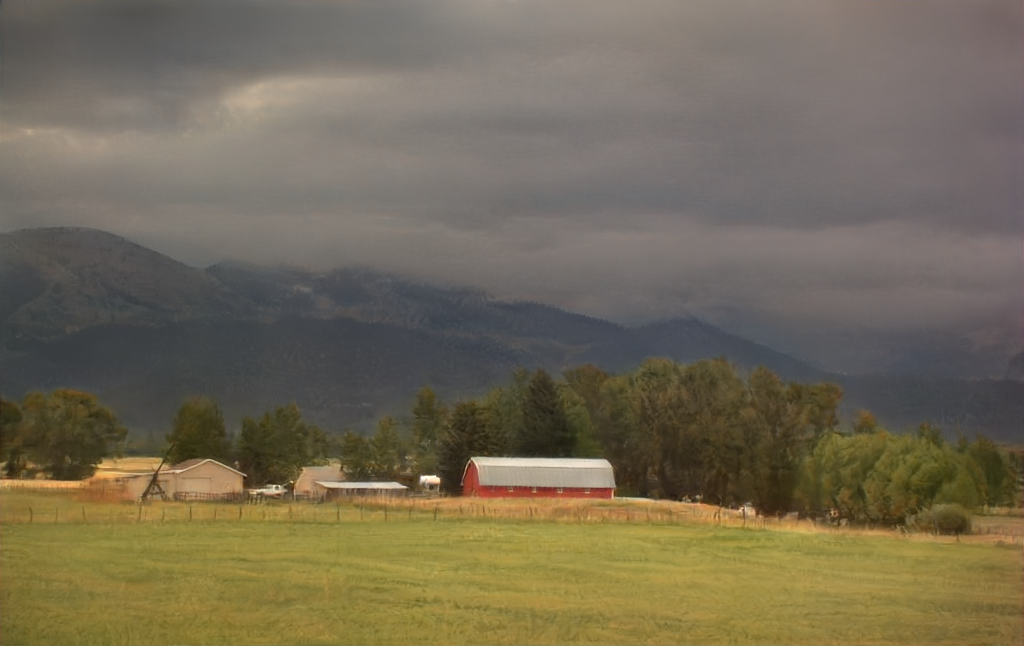}\\%
    \rotatebox[origin=l]{90}{\makebox[0mm][l]{\hspace{4mm}8.7.2016\hspace{1.5mm}\footnotesize{12:00}}}\hfill%
    \includegraphics[width=\h]{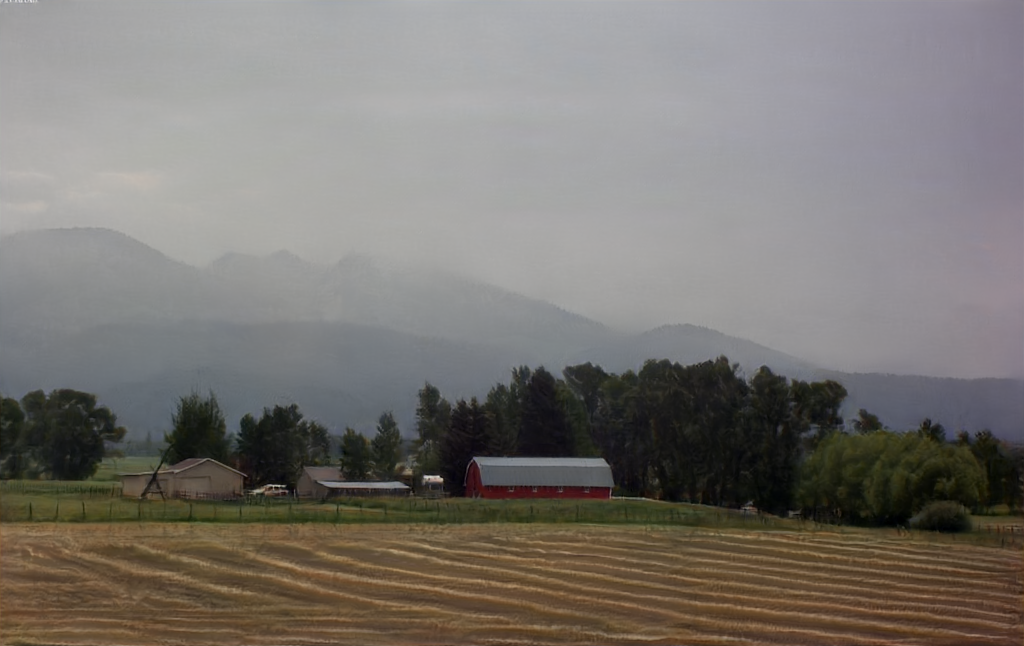}\hfill%
    \includegraphics[width=\h]{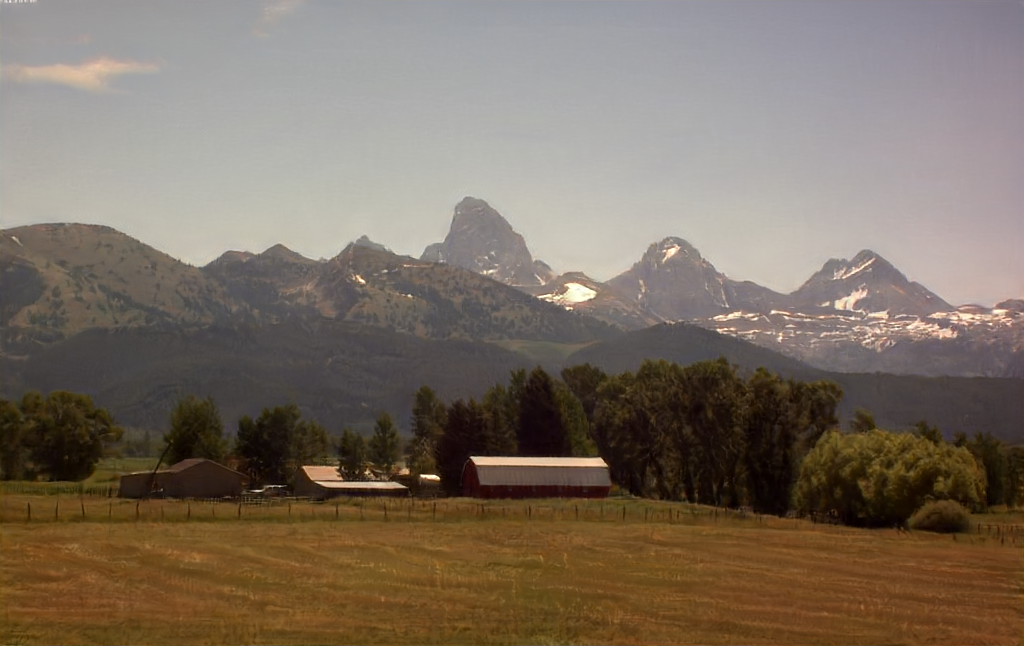}\hfill%
    \includegraphics[width=\h]{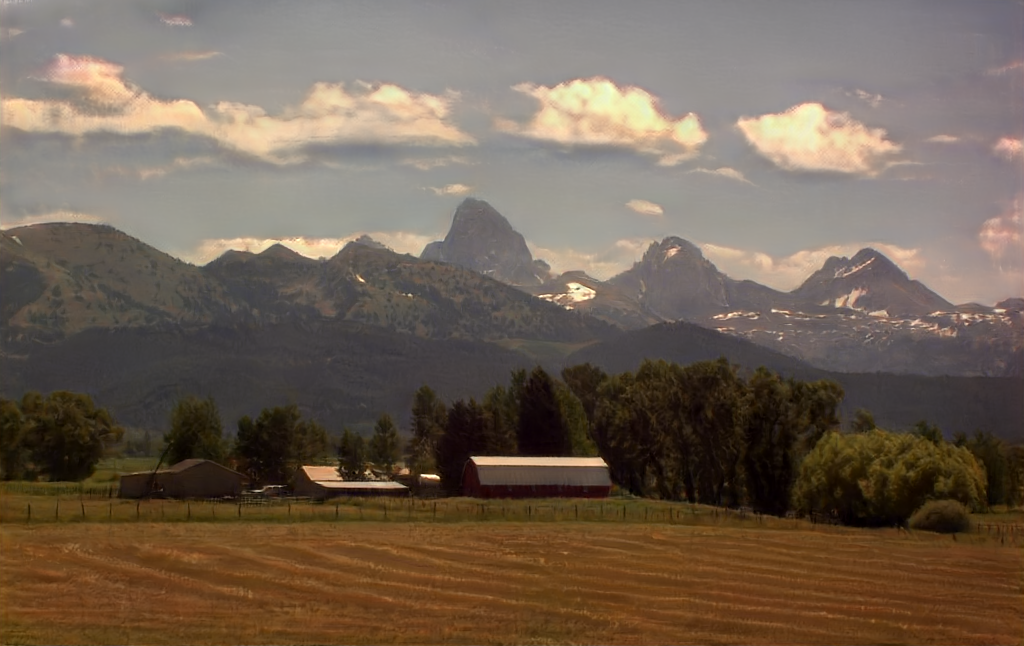}\hfill%
    \includegraphics[width=\h]{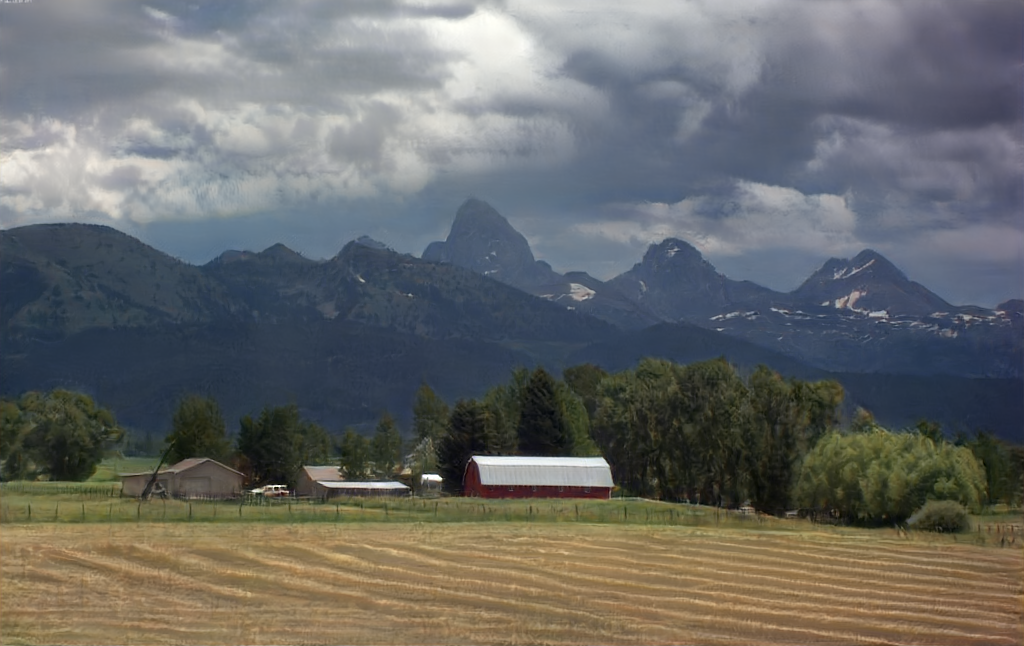}\\%
    \rotatebox[origin=l]{90}{\makebox[0mm][l]{\hspace{4mm}9.2.2017\hspace{1.5mm}\footnotesize{15:00}}}\hfill%
    \includegraphics[width=\h]{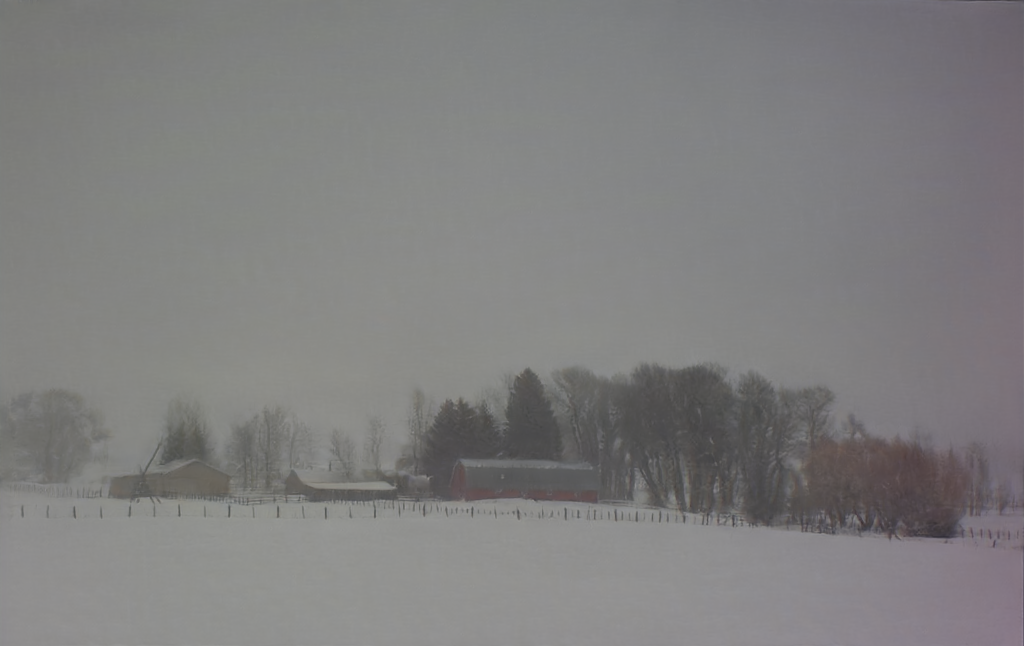}\hfill%
    \includegraphics[width=\h]{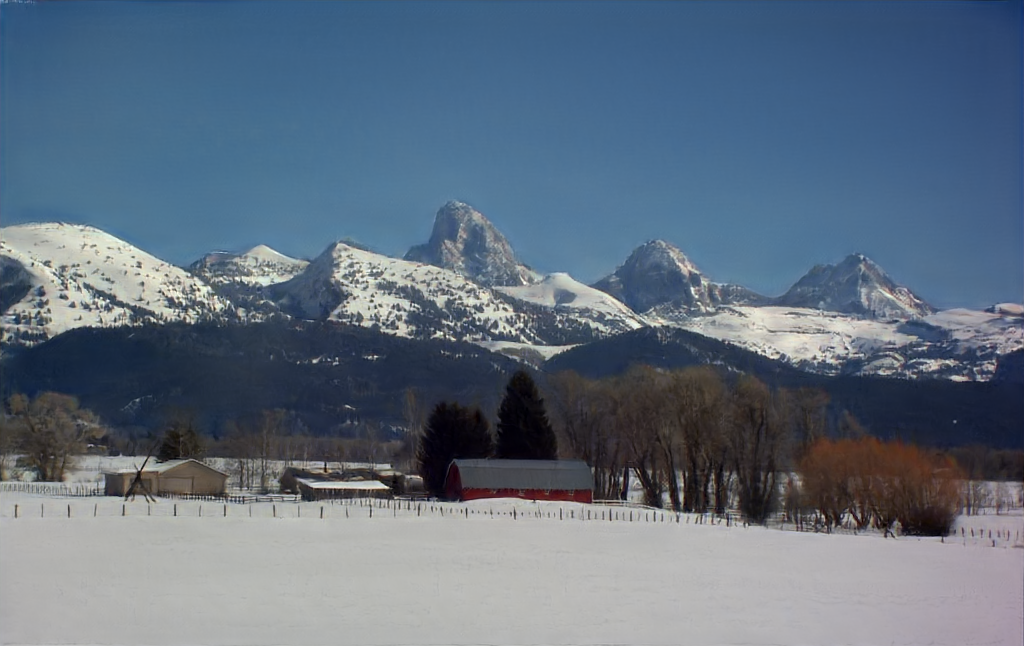}\hfill%
    \includegraphics[width=\h]{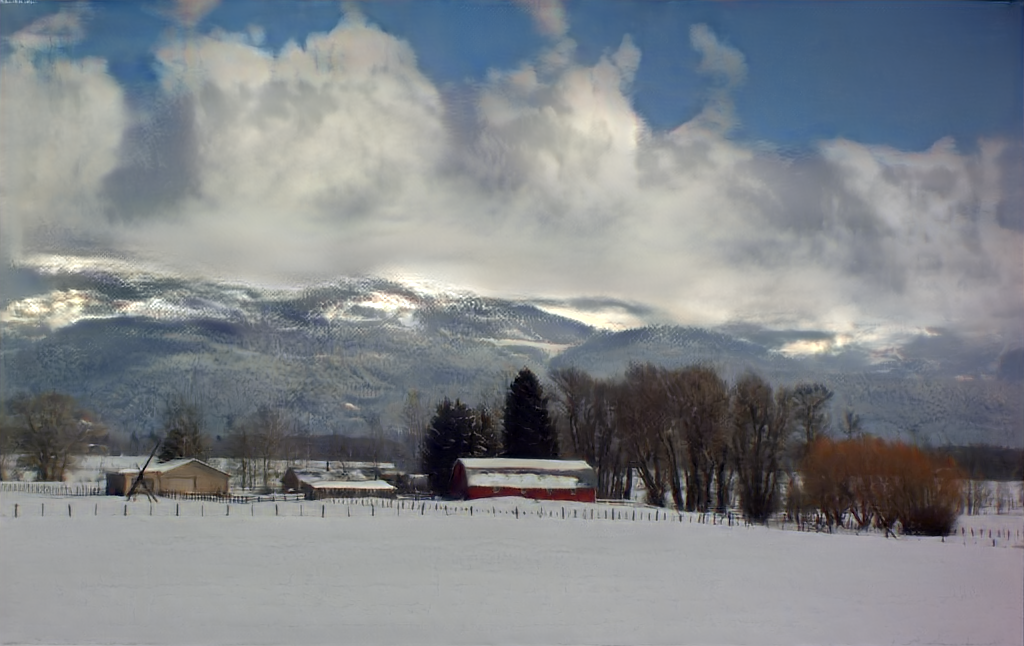}\hfill%
    \includegraphics[width=\h]{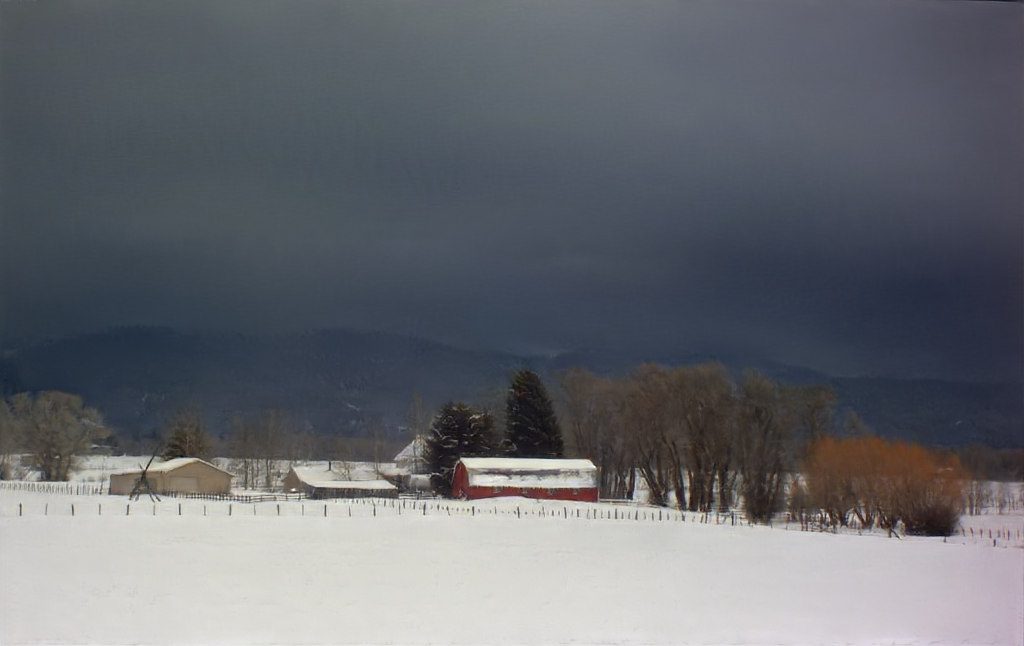}\\%
    \rotatebox[origin=l]{90}{\makebox[0mm][l]{  }}\hfill%
    \makebox[\h]{Latent code 1 (foggy day)}\hfill%
    \makebox[\h]{Latent code 2 (bright day)}\hfill%
    \makebox[\h]{Latent code 3 (partly cloudy)}\hfill%
    \makebox[\h]{Latent code 4 (rainy day)}
    \caption{We synthesize a frame from \Barn{} dataset at four different timestamps (rows), using four latent codes (columns). The latent codes express the weather consistently across timestamps. 
    On each row, all differences between the images are caused by the latent codes.}
    \label{#1}
\end{figure*}
}

\newcommand{\figDefects}[1]{
\renewcommand{\h}{0.62\columnwidth}%
\renewcommand{\hh}{0.33\columnwidth}%
\begin{figure}
    \rotatebox[origin=l]{90}{\makebox[0mm][l]{\normalsize{\phantom{X}}}}\hfill%
    \hdrbox{\h}{7 years}\hfill\makebox[\hh]{}\\
    \rotatebox[origin=l]{90}{\makebox[0mm][l]{\hspace*{0.07\linewidth}\normalsize{(a) \Normandy{}}}}\hfill%
    \includegraphics[trim=0 60 0 60,clip,width=\h]{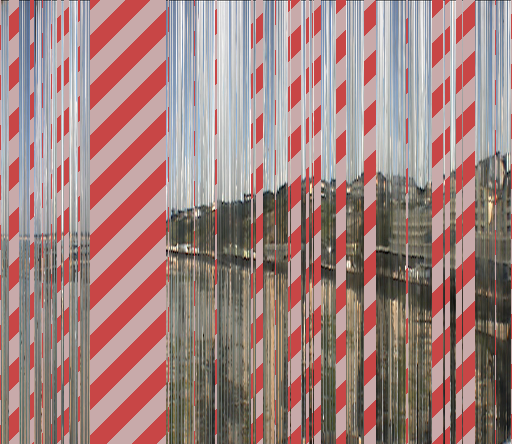}\hfill%
    \includegraphics[trim=0 25 0 25,clip,width=\hh]{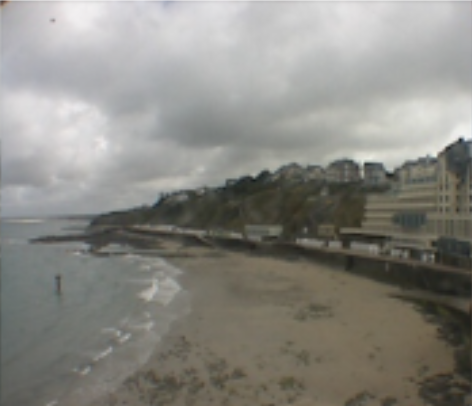}\\%
    \rotatebox[origin=l]{90}{\makebox[0mm][l]{\normalsize{\phantom{X}}}}\hfill%
    \hdrbox{\h}{5 years}\hfill\makebox[\hh]{}\\
    \rotatebox[origin=l]{90}{\makebox[0mm][l]{\hspace*{0.09\linewidth}\normalsize{\hspace{1mm}(b) \Barn{}}}}\hfill%
    \includegraphics[trim=0 10 0 10,clip,width=\h]{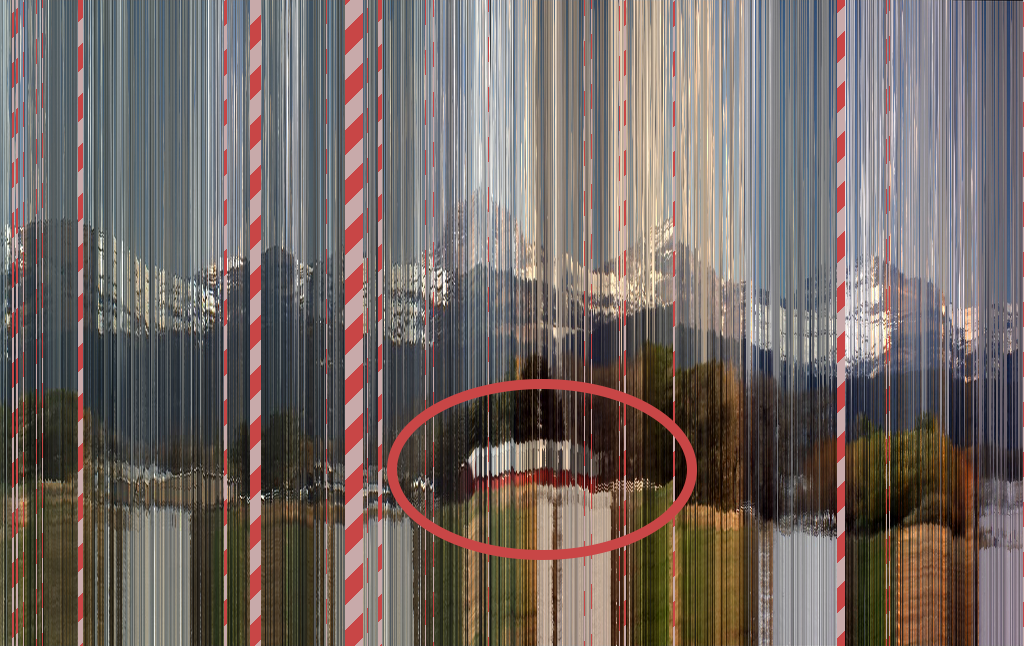}\hfill%
    \includegraphics[trim=0 0 0 0,clip,width=\hh]{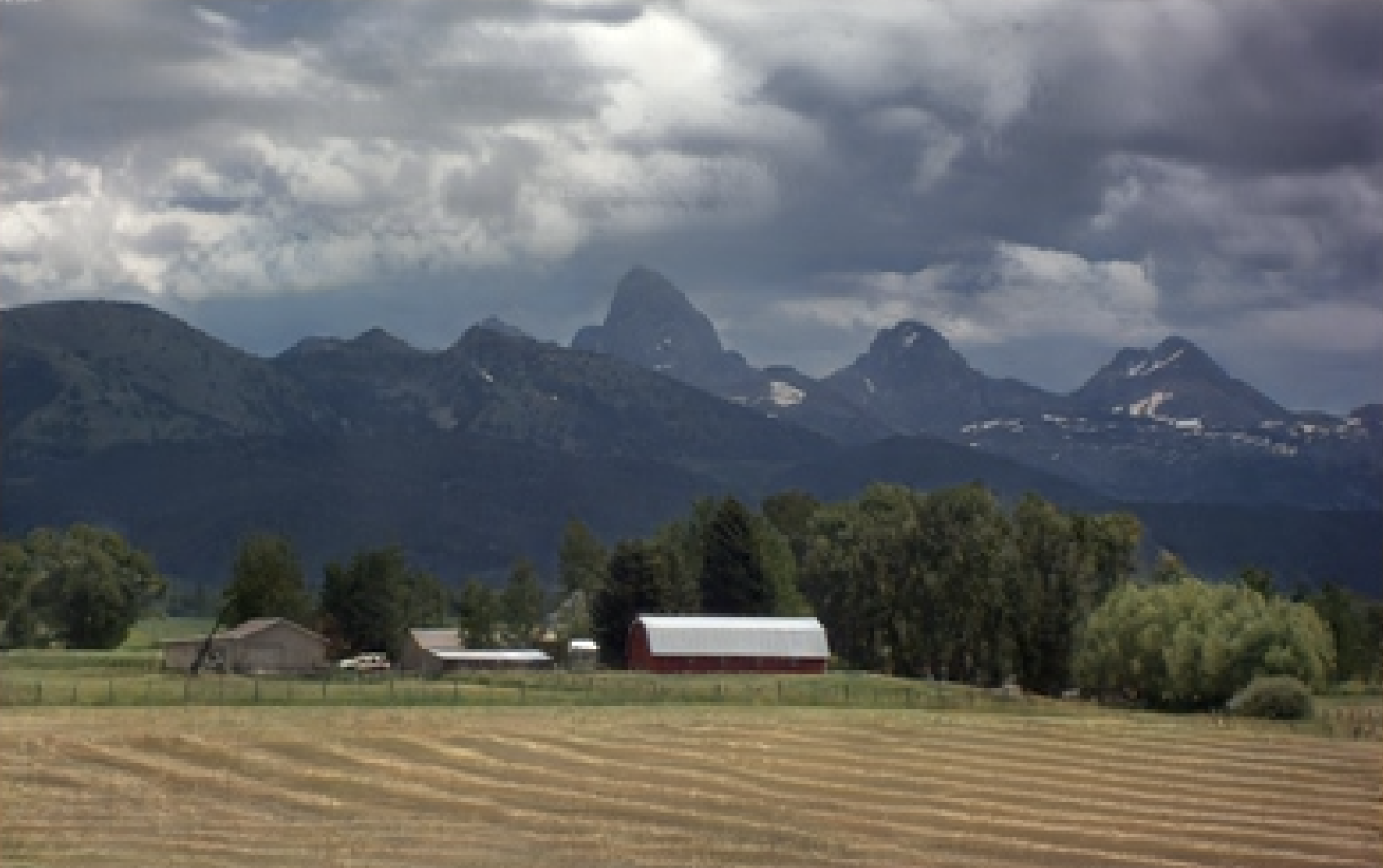}\\%
    \rotatebox[origin=l]{90}{\makebox[0mm][l]{\normalsize{}}}\hfill%
    \makebox[\h]{Time-lapse image of original dataset}\hfill%
    \makebox[\hh]{Example frame}
    \caption{Time-lapse datasets contain various defects. (a) In this dataset long chunks of data are missing (striped columns).  (b) Camera alignment has changed in the middle, causing the barn to bend upwards.}
    \label{#1}
\end{figure}
}

\newcommand{\figTimelapseImage}[1]{
\renewcommand{\h}{0.4\columnwidth}%
\renewcommand{\hh}{0.4\columnwidth}%
\begin{figure}
    \centering
    \includegraphics[width=\columnwidth]{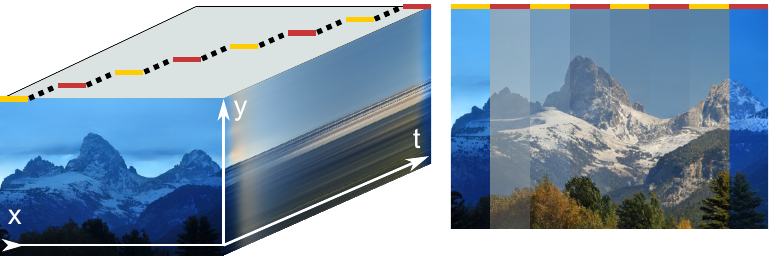}
    \makebox[\h]{Time-lapse data (3D)}\hfill%
    \makebox[\hh]{Time-lapse image}
    \caption{A \emph{time-lapse image} shows a path through the three-dimensional time-lapse (left) in a single image by stacking vertical spans from individual images together (right). The sources of the spans within the time-lapse cube have been illustrated by the lines of alternating color.}
    \label{#1}
\end{figure}
}

\newcommand{\figAblationFcat}[1]{
\renewcommand{\h}{\linewidth}%
\begin{figure*}
    \includegraphics[width=\h]{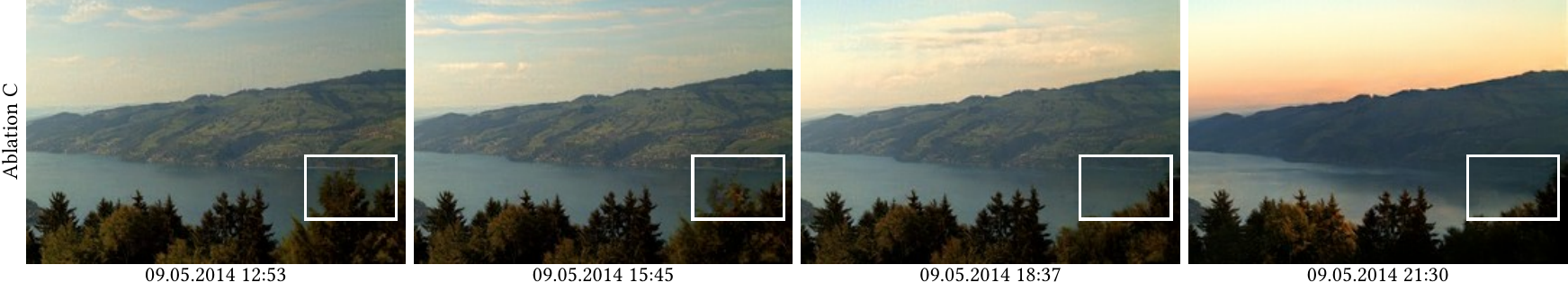}\\%
    \vspace{-0.5\baselineskip}
    \caption{\radd{With the full stack of cyclic conditioning signals the concatenation conditioning mechanism of Mirza and Osindero~\shortcite{Mirza2014} (Ablation C) tends to entangle the inputs, here seen as the day cycle controlling changes in the image that we would expect to be \mbox{handled entirely by the trend input (a tree being cut down).}}}
    \label{#1}
\end{figure*}
}

\newcommand{\figAblationSingleCond}[1]{
\renewcommand{\h}{\linewidth}%
\begin{figure*}
    \includegraphics[width=\h]{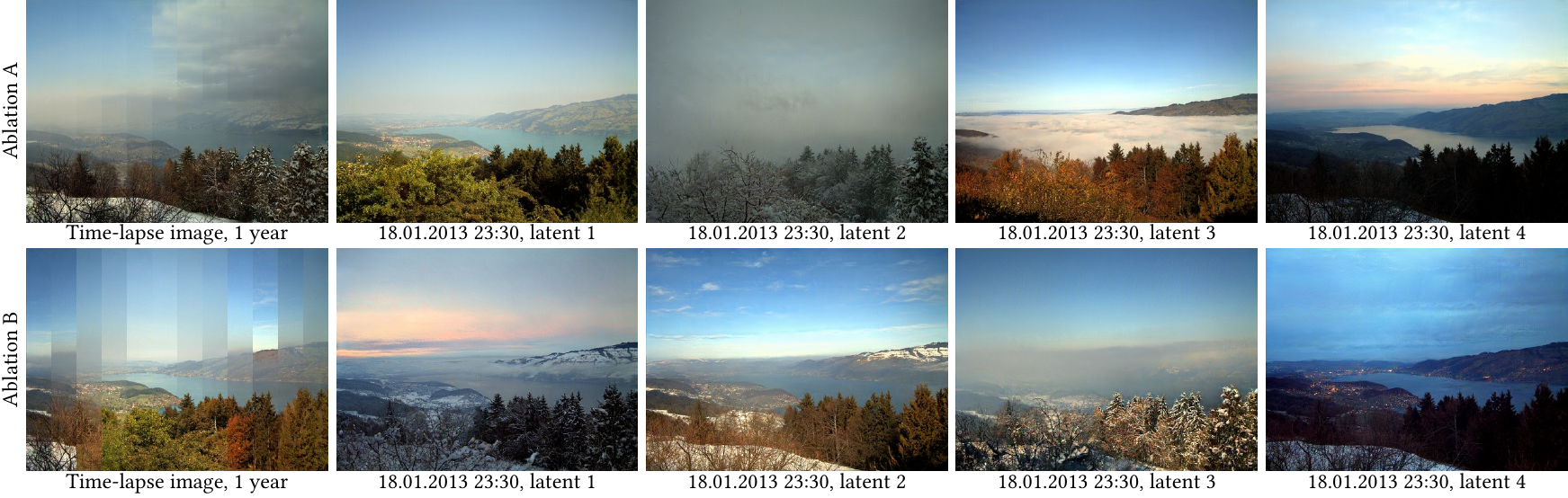}\\%
    \vspace{-0.5\baselineskip}
    \caption{\radd{Two ablations with a single scalar time input. Ablation A: with our proposed style modulation conditioning mechanism, the input controls only trends, with all other variation ending up in the latent space. This results in seemingly broken time-lapse images that display only trend changes, and random samples that have completely random time of day and season. Ablation B: using the concatenation method of Mirza and Osindero~\shortcite{Mirza2014}, the conditioning ends up controlling trends and time of year together, with the latent space controlling the day cycle. This results in random samples that are consistent w.r.t. the year cycle, but have inconsistent time of day.}}%
    \label{#1}
\end{figure*}
}

\newcommand{\figTrend}[1]{
\renewcommand{\h}{0.29\textwidth}%
\renewcommand{\hh}{0.38\textwidth}%
\renewcommand{\hhh}{0.981\textwidth}
\begin{figure*}
    \rotatebox[origin=l]{90}{\makebox[0mm][l]{\hspace*{10mm}\Frankfurt{}}}\hfill%
    \includegraphics[width=\hhh]{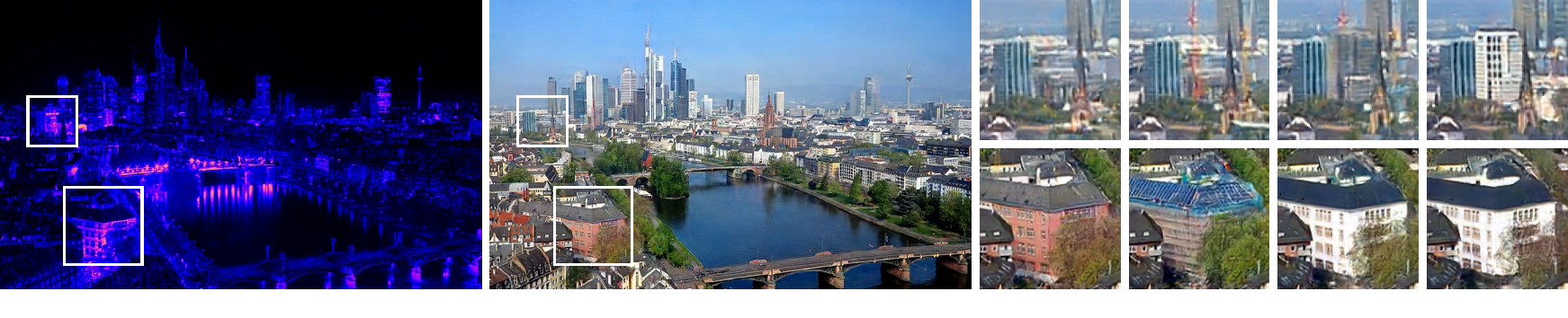}\\
    \rotatebox[origin=l]{90}{\makebox[0mm][l]{\hspace*{7mm}\MPP{}}}\hfill%
    \includegraphics[width=\hhh]{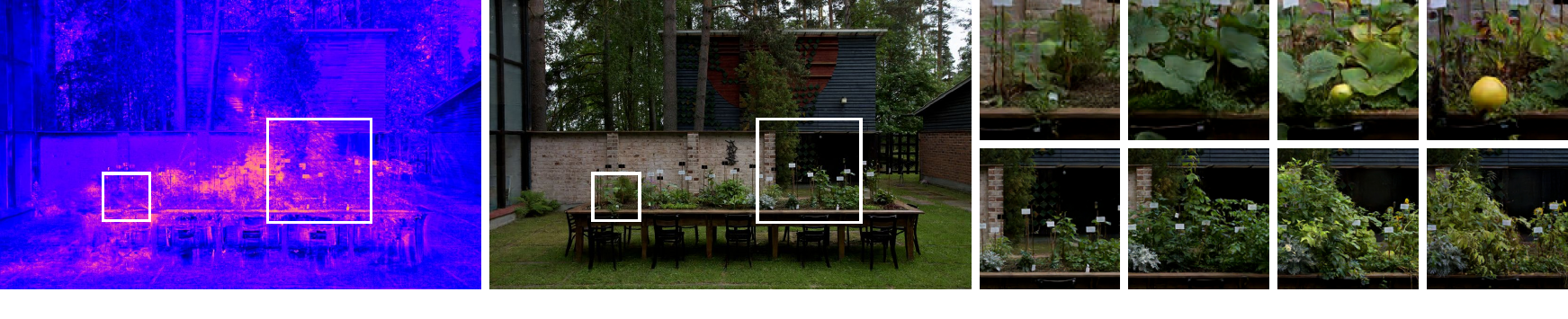}\\
    \rotatebox[origin=l]{90}{\makebox[0mm][l]{\hspace*{10mm}\Teton{}}}\hfill%
    \includegraphics[width=\hhh]{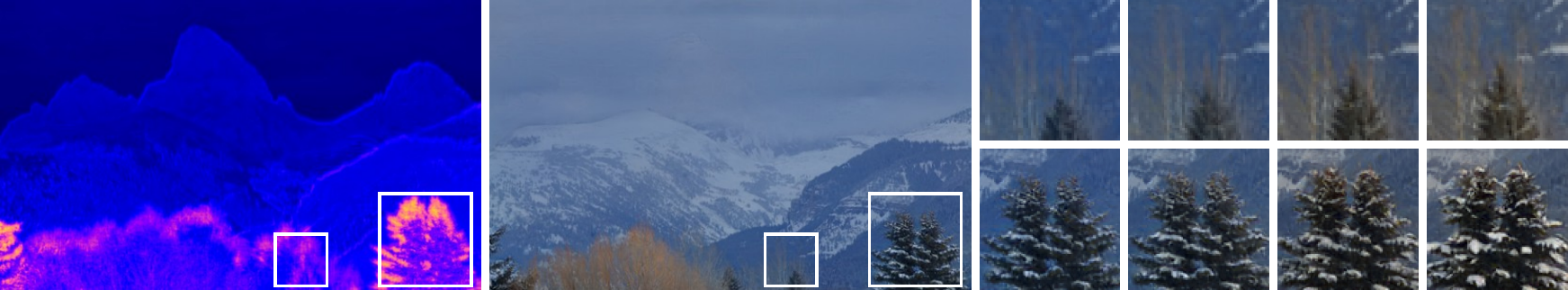}\\
    \hspace{5mm}%
    \makebox[\h]{Variance from the trend signal}\hfill%
    \makebox[\h]{Example image}\hfill%
    \makebox[\hh]{Evolution of the close-ups over time}\\
\caption{The trend signal ends up controlling mainly plant growth, and construction and renovation of buildings. The variance images highlight the areas most strongly affected by the trend signal. Note that in \Frankfurt{} lights were added to the upper bridge in the middle of the time-lapse sequence, and thus they show up in the trend signal. }%
\vspace{2\baselineskip}   
    \label{#1}
\end{figure*}
}

\newcommand{\figComparison}[1]{
\renewcommand{\h}{0.353\textwidth}%
\renewcommand{\hh}{0.149\textwidth}%
\begin{figure*}
    \centering
    \includegraphics[width=\h]{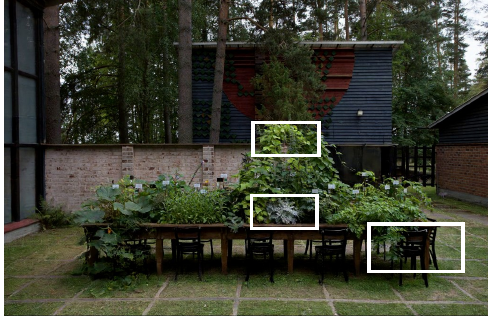}%
    \includegraphics[width=\hh]{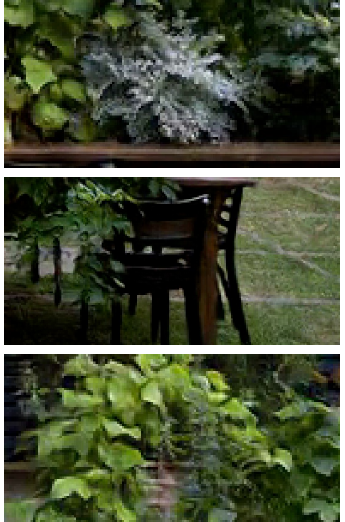}%
    \includegraphics[width=\hh]{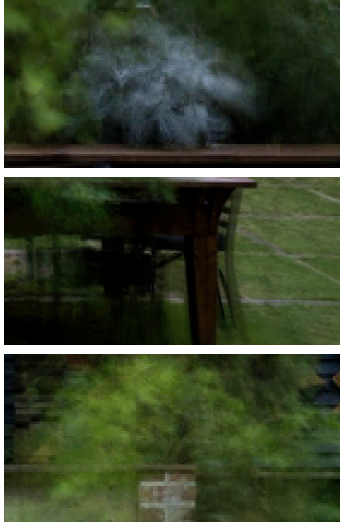}%
    \includegraphics[width=\h]{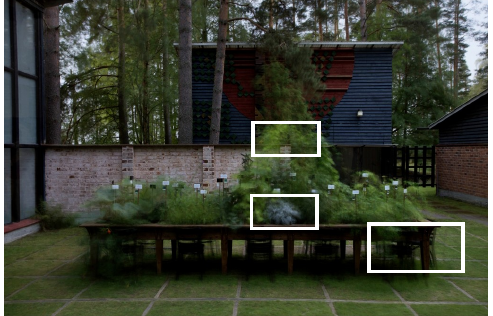}\\
    \makebox[\h]{Ours}\hfill%
    \makebox[0.134\textwidth]{Closeups}\hfill%
    \makebox[\h]{Martin-Brualla et al.}\\%
    \caption{Comparison of our result with \cite{MartinBrualla2015TimelapseMining} in \MPP{}. Our method gives significantly sharper results in case of moving content such as growing plants or the moving chairs.}%
    \label{#1}
\end{figure*}
}

\newcommand{\figRobustness}[1]{
\renewcommand{\h}{0.495\columnwidth}%
\begin{figure}
    \hdrbox{\h}{12 months}\hfill%
    \hdrbox{\h}{12 months}\\
    \includegraphics[trim=0 0 0 0,clip,width=\h]{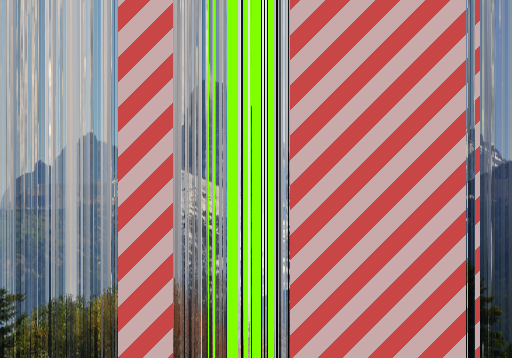}\hfill%
    \includegraphics[trim=0 0 0 0,clip,width=\h]{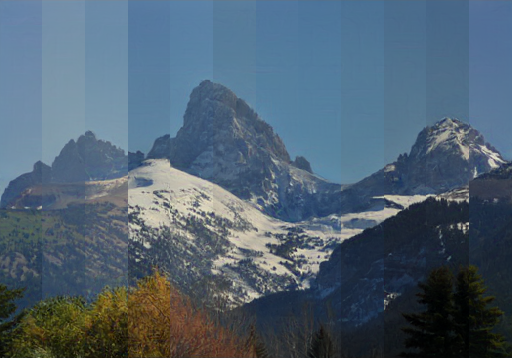}\\
    \makebox[\h]{Input data}\hfill%
    \makebox[\h]{Our example result}
    \caption{Almost half of the input data is missing from \Teton{} in 2016, marked with red stripes. Furthermore, exposure problems have rendered many days unusable (yellow-green column). We can synthesize plausible time-lapses even in such challenging cases.}
    \label{#1}
\end{figure}
}

\newcommand{\tabSceneDetails}[1]{
\begin{table*}
\renewcommand{\s}{\hspace*{1.5mm}}
\renewcommand{\ss}{\hspace*{2.2mm}}
\renewcommand{\sss}{\hspace*{2.5mm}}
\renewcommand{\arraystretch}{1.15}
\vspace*{-0.55\baselineskip}
\caption{Details of our test datasets.}
\vspace*{-0.25\baselineskip}
\begin{tabular}{ |>{\raggedright}p{2.05cm}|>{\centering}p{1.45cm}|>{\centering}p{1.05cm}|>{\centering}p{2.25cm}|>{\centering}p{1.0cm}|>{\centering}p{2.0cm}|>{\centering}p{1.8cm}|p{1.3cm}||p{1.7cm}| }
\hline
\textbf{Name} & \textbf{Resolution} & \textbf{Frames} & \textbf{Date range} & \textbf{Length} & \textbf{Sampling rate} & \textbf{Longest gaps} & \textbf{AMOS ID} & \textbf{Trend noise}\\
 & & & & \footnotesize{(days)} & \footnotesize{(median, 95th perc.)} & \footnotesize{(days)} & & \hspace{6mm}\footnotesize{($\sigma_g$)} \\
 \hline
\Barn{} & $1024\times646$ & \s63806 & 05.2012 -- 02.2017 & 1729 & 0.50h, 0.98h & 32, 18, 13 & \s\s19189 & \sss1.5 years\\
\Frankfurt{} & $1024\times960$ & \s52118 & 04.2013 -- 07.2016 & 1197 & 0.50h, 0.99h & 11, 5, 4 & \s\s09483 & \sss1.5 years \\
\Kuessnacht{} & \s$512\times358$ & \s54966 & 10.2012 -- 06.2017 & 1686 & 0.50h, 0.99h & 15, 8, 5 & \s\s08687 & \sss1.5 years \\
\Muotathal{} & $1024\times786$ & \s82374 & 02.2010 -- 05.2017 & 2661 & 0.30h, 2.88h & 34, 26, 13 & \s\s10180 & \sss\ss2 years \\
\Normandy{} & \s$512\times444$ & \s56247 & 02.2010 -- 10.2016 & 2440 & 0.50h, 1.01h & 365, 59, 58 & \s\s09780 & \sss\ss2 years \\
\Teton{} & \s$512\times358$ & \s48276 & 05.2012 -- 06.2017 & 1841 & 0.50h, 2.00h & 48, 39, 37 & \s\s19188 & \sss1.5 years \\
\Twomedicine{} & $1008\times536$ & \s79594 & 08.2011 -- 05.2017 & 2120 & 0.50h, 0.99h & 112, 69, 43 & \s\s17183 & \sss\ss2 years \\
\Valley{} & $1024\times828$ & \s49834 & 03.2011 -- 04.2017 & 2225 & 0.50h, 1.00h & 15, 13, 8 & \s\s07371 & \sss1.5 years \\
\hline
\MPP{} & $1024\times670$ & \s\s6338 & 05.2019 -- 10.2019 & \s143 & 0.50h, 0.51h & 8, 0, 0 & \s\s\s\s--- & \sss1 week\\
\hline
\end{tabular}
\vspace*{-0.65\baselineskip}
\label{#1}
\end{table*}
}

\newcommand{\tabFIDcomp}[1]{
\begin{table*}
\caption{FIDs from equal-time training runs of our method and StyleGAN2 (unconditional). Despite the increased control provided by our method, we do not see a systematic decrease in image quality.}%
\begin{tabular} { |m{2.0cm}|m{1.4cm}|m{2.15cm}|m{1.4cm}|m{1.0cm}|m{0.8cm}|m{0.9cm}|m{1.4cm}|m{1.3cm}|m{1.53cm}| }\hline
Method & \Kuessnacht{} & \MPP{} & \Frankfurt{} & \Valley{} & \Barn{} & \Teton{} & \Normandy{} & \Twomedicine{} & \Muotathal{} \\
\hline Ours & 8.71 & \textbf{9.02} & \textbf{11.11} & \textbf{3.83} & \textbf{4.66} & 6.88 & \textbf{3.51} & 4.76 & \textbf{5.76}\\
StyleGAN2 & \textbf{8.13} & 11.44 & 11.65 & 3.87 & 6.83 & \textbf{6.21} & 4.96 & \textbf{4.42} & 6.86\\
\hhline{==========}
Training time & 30h & 60h & 60h & 60h & 60h & 30h & 30h & 60h & 60h\\
\hline
\end{tabular}
\label{#1}
\end{table*}
}



%% file: figures/convergence/conv_out.pdf_tex
\begingroup%
  \makeatletter%
  \providecommand\color[2][]{%
    \errmessage{(Inkscape) Color is used for the text in Inkscape, but the package 'color.sty' is not loaded}%
    \renewcommand\color[2][]{}%
  }%
  \providecommand\transparent[1]{%
    \errmessage{(Inkscape) Transparency is used (non-zero) for the text in Inkscape, but the package 'transparent.sty' is not loaded}%
    \renewcommand\transparent[1]{}%
  }%
  \providecommand\rotatebox[2]{#2}%
  \newcommand*\fsize{\dimexpr\f@size pt\relax}%
  \newcommand*\lineheight[1]{\fontsize{\fsize}{#1\fsize}\selectfont}%
  \ifx\svgwidth\undefined%
    \setlength{\unitlength}{460.8bp}%
    \ifx\svgscale\undefined%
      \relax%
    \else%
      \setlength{\unitlength}{\unitlength * \real{\svgscale}}%
    \fi%
  \else%
    \setlength{\unitlength}{\svgwidth}%
  \fi%
  \global\let\svgwidth\undefined%
  \global\let\svgscale\undefined%
  \makeatother%
  \begin{picture}(1,0.575)%
    \lineheight{1}%
    \setlength\tabcolsep{0pt}%
    \put(0,0){\includegraphics[width=\unitlength,page=1]{figures/convergence/conv_out.pdf}}%
  \end{picture}%
\endgroup%

%% file: figures/collapse/mem_out.pdf_tex
\begingroup%
  \makeatletter%
  \providecommand\color[2][]{%
    \errmessage{(Inkscape) Color is used for the text in Inkscape, but the package 'color.sty' is not loaded}%
    \renewcommand\color[2][]{}%
  }%
  \providecommand\transparent[1]{%
    \errmessage{(Inkscape) Transparency is used (non-zero) for the text in Inkscape, but the package 'transparent.sty' is not loaded}%
    \renewcommand\transparent[1]{}%
  }%
  \providecommand\rotatebox[2]{#2}%
  \newcommand*\fsize{\dimexpr\f@size pt\relax}%
  \newcommand*\lineheight[1]{\fontsize{\fsize}{#1\fsize}\selectfont}%
  \ifx\svgwidth\undefined%
    \setlength{\unitlength}{460.8bp}%
    \ifx\svgscale\undefined%
      \relax%
    \else%
      \setlength{\unitlength}{\unitlength * \real{\svgscale}}%
    \fi%
  \else%
    \setlength{\unitlength}{\svgwidth}%
  \fi%
  \global\let\svgwidth\undefined%
  \global\let\svgscale\undefined%
  \makeatother%
  \begin{picture}(1,0.575)%
    \lineheight{1}%
    \setlength\tabcolsep{0pt}%
    \put(0,0){\includegraphics[width=\unitlength,page=1]{figures/collapse/mem_out.pdf}}%
  \end{picture}%
\endgroup%

%% file: main.bbl
\begin{thebibliography}{}

\bibitem[Anokhin et~al., 2020]{Anokhin2020HiDT}
Anokhin, I., Solovev, P., Korzhenkov, D., Kharlamov, A., Khakhulin, T.,
  Silvestrov, A., Nikolenko, S., Lempitsky, V., and Sterkin, G. (2020).
\newblock High-resolution daytime translation without domain labels.
\newblock In {\em Proc. CVPR}.

\bibitem[Brock et~al., 2019]{Brock2019BigGAN}
Brock, A., Donahue, J., and Simonyan, K. (2019).
\newblock Large scale gan training for high fidelity natural image synthesis.
\newblock In {\em Proc. ICLR}.

\bibitem[Choi et~al., 2020]{Choi2020StarGAN2}
Choi, Y., Uh, Y., Yoo, J., and Ha, J.-W. (2020).
\newblock Stargan v2: Diverse image synthesis for multiple domains.
\newblock In {\em Proc. CVPR}.

\bibitem[Chong et~al., 2021]{Chong2021}
Chong, M.~J., Chu, W.-S., Kumar, A., and Forsyth, D. (2021).
\newblock Retrieve in style: Unsupervised facial feature transfer and
  retrieval.
\newblock In {\em Proc. ICCV}.

\bibitem[Clark et~al., 2019]{Clark2019DVDGAN}
Clark, A., Donahue, J., and Simonyan, K. (2019).
\newblock Efficient video generation on complex datasets.
\newblock {\em CoRR}, abs/1907.06571.

\bibitem[Collins et~al., 2020]{Collins2020}
Collins, E., Bala, R., Price, B., and S{\"u}sstrunk, S. (2020).
\newblock Editing in style: Uncovering the local semantics of {GANs}.
\newblock In {\em Proc. CVPR}.

\bibitem[Colton and Ferrer, 2021]{Colton2021GANlapse}
Colton, S. and Ferrer, B.~P. (2021).
\newblock Ganlapse generative photography.
\newblock In {\em Proc. International Conference on Computational Creativity}.

\bibitem[Dinh et~al., 2017]{Dinh2016RealNVP}
Dinh, L., Sohl-Dickstein, J., and Bengio, S. (2017).
\newblock Density estimation using {R}eal {NVP}.
\newblock In {\em Proc. ICLR}.

\bibitem[Endo et~al., 2019]{Endo2019TimelapseWarp}
Endo, Y., Kanamori, Y., and Kuriyama, S. (2019).
\newblock Animating landscape: self-supervised learning of decoupled motion and
  appearance for single-image video synthesis.
\newblock In {\em Proc. SIGGRAPH ASIA 2019}.

\bibitem[Goodfellow et~al., 2014]{Goodfellow2014GAN}
Goodfellow, I., Pouget-Abadie, J., Mirza, M., Xu, B., Warde-Farley, D., Ozair,
  S., Courville, A., and Bengio, Y. (2014).
\newblock {Generative Adversarial Networks}.
\newblock In {\em Proc. NIPS}.

\bibitem[H\"{a}rk\"{o}nen et~al., 2020]{Harkonen2019GANSpace}
H\"{a}rk\"{o}nen, E., Hertzmann, A., Lehtinen, J., and Paris, S. (2020).
\newblock {GANSpace}: Discovering interpretable {GAN} controls.
\newblock In {\em Proc. NeurIPS}.

\bibitem[Heusel et~al., 2017]{Heusel2017FID}
Heusel, M., Ramsauer, H., Unterthiner, T., Nessler, B., and Hochreiter, S.
  (2017).
\newblock {GAN}s trained by a two time-scale update rule converge to a local
  nash equilibrium.
\newblock In {\em Proceedings of the 31st International Conference on Neural
  Information Processing Systems}, NIPS'17, page 6629–6640, Red Hook, NY,
  USA. Curran Associates Inc.

\bibitem[Ho et~al., 2020]{Ho2020DDPM}
Ho, J., Jain, A., and Abbeel, P. (2020).
\newblock Denoising diffusion probabilistic models.
\newblock In {\em Proc. NeurIPS}.

\bibitem[Horita and Yanai, 2020]{Daichi2020SSAGAN}
Horita, D. and Yanai, K. (2020).
\newblock Ssa-gan: End-to-end time-lapse video generation with spatial
  self-attention.
\newblock In {\em Proc. ACPR}.

\bibitem[Huang et~al., 2021]{Huang2021PoEGAN}
Huang, X., Mallya, A., Wang, T.-C., and Liu, M.-Y. (2021).
\newblock Multimodal conditional image synthesis with product-of-experts
  {GANs}.
\newblock {\em CoRR}, abs/2112.05130.

\bibitem[Jacobs et~al., 2009]{jacobs09webcamgis}
Jacobs, N., Burgin, W., Fridrich, N., Abrams, A., Miskell, K., Braswell, B.~H.,
  Richardson, A.~D., and Pless, R. (2009).
\newblock The global network of outdoor webcams: Properties and applications.
\newblock In {\em ACM SIGSPATIAL International Conference on Advances in
  Geographic Information Systems (ACM SIGSPATIAL)}.

\bibitem[Jacobs et~al., 2007]{Jacobs07AMOS}
Jacobs, N., Roman, N., and Pless, R. (2007).
\newblock Consistent temporal variations in many outdoor scenes.
\newblock In {\em Proc. CVPR}.

\bibitem[Kafri et~al., 2021]{Kafri2021}
Kafri, O., Patashnik, O., Alaluf, Y., and Cohen{-}Or, D. (2021).
\newblock Stylefusion: {A} generative model for disentangling spatial segments.
\newblock {\em CoRR}, abs/2107.07437.

\bibitem[{Karacan} et~al., 2019]{Karacan2019GauGANish}
{Karacan}, L., {Akata}, Z., {Erdem}, A., and {Erdem}, E. (2019).
\newblock Manipulating attributes of natural scenes via hallucination.
\newblock In {\em Proc. TOG}.

\bibitem[Karras et~al., 2020a]{Karras2020ADA}
Karras, T., Aittala, M., Hellsten, J., Laine, S., Lehtinen, J., and Aila, T.
  (2020a).
\newblock Training generative adversarial networks with limited data.
\newblock In {\em Proc. NeurIPS}.

\bibitem[Karras et~al., 2021]{Karras2021AliasFree}
Karras, T., Aittala, M., Laine, S., H\"ark\"onen, E., Hellsten, J., Lehtinen,
  J., and Aila, T. (2021).
\newblock Alias-free generative adversarial networks.
\newblock In {\em Proc. NeurIPS}.

\bibitem[Karras et~al., 2019]{Karras2019StyleGAN}
Karras, T., Laine, S., and Aila, T. (2019).
\newblock A style-based generator architecture for generative adversarial
  networks.
\newblock In {\em Proc. CVPR}.

\bibitem[Karras et~al., 2020b]{Karras2020stylegan2}
Karras, T., Laine, S., Aittala, M., Hellsten, J., Lehtinen, J., and Aila, T.
  (2020b).
\newblock Analyzing and improving the image quality of {StyleGAN}.
\newblock In {\em Proc. CVPR}.

\bibitem[Kim et~al., 2020]{Kim2020UGATIT}
Kim, J., Kim, M., Kang, H., and Lee, K. (2020).
\newblock U-gat-it: Unsupervised generative attentional networks with adaptive
  layer-instance normalization for image-to-image translation.
\newblock In {\em Proc. ICLR}.

\bibitem[Kingma and Dhariwal, 2018]{Kingma2018Glow}
Kingma, D.~P. and Dhariwal, P. (2018).
\newblock Glow: Generative flow with invertible 1x1 convolutions.
\newblock In {\em Proc. NeurIPS}.

\bibitem[Kingma and Welling, 2014]{Kingma2014VAE}
Kingma, D.~P. and Welling, M. (2014).
\newblock Auto-encoding variational bayes.
\newblock In {\em Proc. ICLR}.

\bibitem[Logacheva et~al., 2020]{Logacheva2020DeepLandscape}
Logacheva, E., Suvorov, R., Khomenko, O., Mashikhin, A., and Lempitsky, V.
  (2020).
\newblock Deeplandscape: Adversarial modeling of landscape videos.
\newblock In {\em Proc. ECCV}.

\bibitem[Martin-Brualla et~al., 2015]{MartinBrualla2015TimelapseMining}
Martin-Brualla, R., Gallup, D., and Seitz, S.~M. (2015).
\newblock Time-lapse mining from internet photos.
\newblock In {\em Proc. TOG}.

\bibitem[Mirza and Osindero, 2014]{Mirza2014}
Mirza, M. and Osindero, S. (2014).
\newblock Conditional generative adversarial nets.
\newblock {\em CoRR}, abs/1411.1784.

\bibitem[Miyato and Koyama, 2018]{Miyato2018ProjD}
Miyato, T. and Koyama, M. (2018).
\newblock cgans with projection discriminator.
\newblock In {\em Proc. ICLR}.

\bibitem[Nam et~al., 2019]{Nam2019VideoSynthesis}
Nam, S., Ma, C., Chai, M., Brendel, W., Xu, N., and Kim, S.~J. (2019).
\newblock End-to-end time-lapse video synthesis from a single outdoor image.
\newblock In {\em Proc. CVPR}.

\bibitem[Park et~al., 2019]{Park2019SPADE}
Park, T., Liu, M.-Y., Wang, T., and Zhu, J.-Y. (2019).
\newblock Semantic image synthesis with spatially-adaptive normalization.
\newblock In {\em Proc. CVPR}.

\bibitem[Park et~al., 2020]{Park2020SwappingAE}
Park, T., Zhu, J.-Y., Wang, O., Lu, J., Shechtman, E., Efros, A.~A., and Zhang,
  R. (2020).
\newblock Swapping autoencoder for deep image manipulation.
\newblock In {\em Proc. NeurIPS}.

\bibitem[Sohl-Dickstein et~al., 2015]{SohlDickstein2015Diffusion}
Sohl-Dickstein, J., Weiss, E.~A., Maheswaranathan, N., and Ganguli, S. (2015).
\newblock Deep unsupervised learning using nonequilibrium thermodynamics.
\newblock In {\em Proc. ICML}.

\bibitem[Song and Ermon, 2019]{Yang2019DataGradients}
Song, Y. and Ermon, S. (2019).
\newblock Generative modeling by estimating gradients of the data distribution.
\newblock In {\em Proc. NeurIPS}.

\bibitem[Sun et~al., 2021]{Sun2021LoFTR}
Sun, J., Shen, Z., Wang, Y., Bao, H., and Zhou, X. (2021).
\newblock {LoFTR}: Detector-free local feature matching with transformers.
\newblock In {\em Proc. CVPR}.

\bibitem[Tancik et~al., 2020]{Tancik2020FourierFeatures}
Tancik, M., Srinivasan, P.~P., Mildenhall, B., Fridovich-Keil, S., Raghavan,
  N., Singhal, U., Ramamoorthi, R., Barron, J.~T., and Ng, R. (2020).
\newblock Fourier features let networks learn high frequency functions in low
  dimensional domains.
\newblock In {\em Proc. NeurIPS}.

\bibitem[Tulyakov et~al., 2018]{Tulyakov2018MoCoGAN}
Tulyakov, S., Liu, M.-Y., Yang, X., and Kautz, J. (2018).
\newblock {MoCoGAN}: Decomposing motion and content for video generation.
\newblock In {\em Proc. CVPR}.

\bibitem[van~den Oord et~al., 2016a]{VanDenOord2016BPixelRNN}
van~den Oord, A., Kalchbrenner, N., and Kavukcuoglu, K. (2016a).
\newblock Pixel recurrent neural networks.
\newblock In {\em Proc. ICML}.

\bibitem[van~den Oord et~al., 2016b]{VanDenOord2016APixelCNN}
van~den Oord, A., Kalchbrenner, N., Vinyals, O., Espeholt, L., Graves, A., and
  Kavukcuoglu, K. (2016b).
\newblock Conditional image generation with {PixelCNN} decoders.
\newblock In {\em Proc. NIPS}.

\bibitem[Vaswani et~al., 2017]{Vaswani2017Attention}
Vaswani, A., Shazeer, N., Parmar, N., Uszkoreit, J., Jones, L., Gomez, A.~N.,
  Kaiser, L., and Polosukhin, I. (2017).
\newblock Attention is all you need.
\newblock In {\em Proc. NeurIPS}.

\bibitem[Wang et~al., 2019]{Wang2019FewshotVid2vid}
Wang, T.-C., Liu, M.-Y., Tao, A., Liu, G., Kautz, J., and Catanzaro, B. (2019).
\newblock Few-shot video-to-video synthesis.
\newblock In {\em Proc. NeurIPS}.

\bibitem[Wang et~al., 2018]{Wang2018Vid2vid}
Wang, T.-C., Liu, M.-Y., Zhu, J.-Y., Liu, G., Tao, A., Kautz, J., and
  Catanzaro, B. (2018).
\newblock Video-to-video synthesis.
\newblock In {\em Proc. NeurIPS}.

\bibitem[Xiong et~al., 2018]{Xiong2018MSDGAN}
Xiong, W., Luo, W., Ma, L., Liu, W., and Luo, J. (2018).
\newblock Learning to generate time-lapse videos using multi-stage dynamic
  generative adversarial networks.
\newblock In {\em Proc. CVPR}.

\bibitem[Zhu et~al., 2017]{Zhu2017CycleGAN}
Zhu, J.-Y., Park, T., Isola, P., and Efros, A.~A. (2017).
\newblock Unpaired image-to-image translation using cycle-consistent
  adversarial networks.
\newblock In {\em Proc. ICCV}.

\end{thebibliography}
